\documentclass[10pt,twocolumn,letterpaper]{article}
\usepackage{iccv}
\usepackage{times}
\usepackage{epsfig}
\usepackage{graphicx}
\usepackage{amsmath}
\usepackage{amssymb}
\usepackage{booktabs}
\usepackage{todonotes}
\usepackage{capt-of}

\usepackage[pagebackref=true,breaklinks=true,letterpaper=true,colorlinks,bookmarks=false]{hyperref}

\iccvfinalcopy 

\begin{document}
	
	\title{MultiStyleGAN: Multiple One-shot Image Stylizations using a Single GAN}
	
	\author{
		Viraj Shah, Ayush Sarkar, Sudharsan Krishnakumar Anitha, Svetlana Lazebnik\\
		University of Illinois\\
		Urbana-Champaign\\
		{\tt\small vjshah3, ayushs2, sk118, slazebni@illinois.edu}
	}
	
	\twocolumn[{%
		\renewcommand\twocolumn[1][]{#1}%
		\maketitle
		\vspace{-2.8em}
		\begin{center}
			\includegraphics[width=0.94\linewidth]{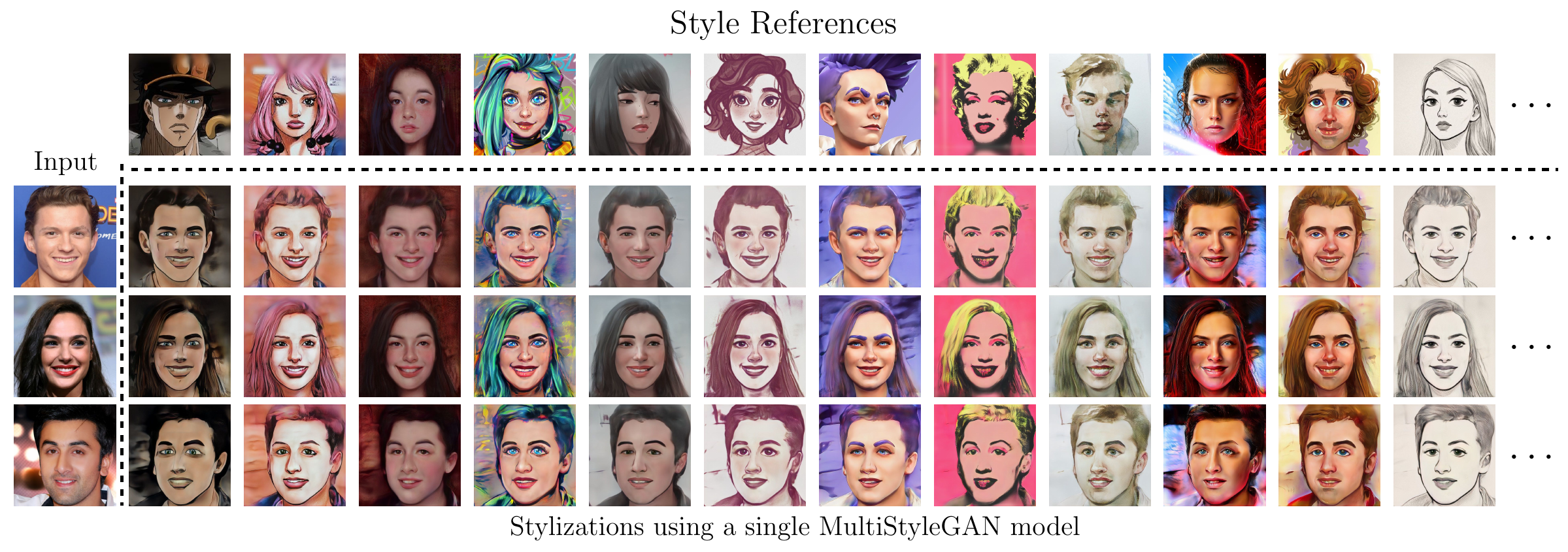}
		\end{center}
		\vspace{-1em}
		\captionof{figure}{\small{We propose MultiStyleGAN for one-shot image stylization that can stylize any input image to multiple reference styles for up to $120$ styles simultaneously. Our method fine-tunes a single pre-trained generator on multiple unique reference styles while requiring only a single image of each style. We show successful stylizations of input images (left column) generated by a single generator fine-tuned on multiple reference styles (top row).
			}
		}
		\label{fig:banner1}
		\vspace{1em}
	}]
	
	\maketitle

	\begin{abstract}
		\vspace{-0.8em}
		Image stylization aims at applying a reference style to arbitrary input images. A common scenario is one-shot stylization, where only one example is available for each reference style. Recent approaches for one-shot stylization such as JoJoGAN~\cite{jojogan} fine-tune a pre-trained StyleGAN2 generator on a single style reference image. However, such methods cannot generate multiple stylizations without fine-tuning a new model for each style separately. In this work, we present a \emph{MultiStyleGAN} method that is capable of producing multiple different stylizations at once by fine-tuning a \emph{single} generator. The key component of our method is a learnable transformation module called Style Transformation Network. It takes latent codes as input, and learns linear mappings to different regions of the latent space to produce distinct codes for each style, resulting in a \emph{multistyle space}. Our model inherently mitigates overfitting since it is trained on multiple styles, hence improving the quality of stylizations. Our method can learn upwards of $120$ image stylizations at once, bringing $8\times$ to $60\times$ improvement in training time over recent competing methods. We support our results through user studies and quantitative results that indicate meaningful improvements over existing methods. 
	\end{abstract}

	\vspace{-1.5em}
	\section{Introduction}
	\label{sec:intro}
	\begin{figure*}
		\begin{center}
			\includegraphics[width=0.90\linewidth]{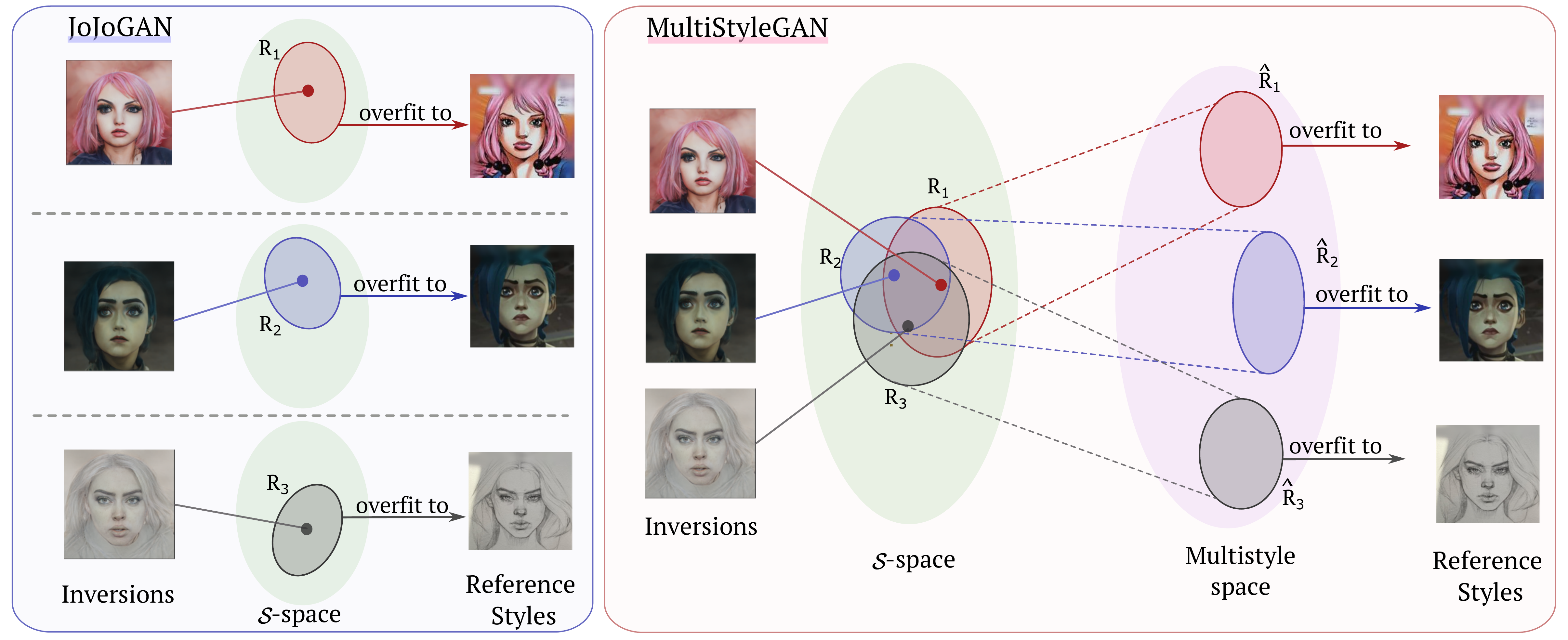}
		\end{center}
		\caption{\small{
				Both JoJoGAN and MultiStyleGAN invert the style reference and apply style-mixing on the inverted code to obtain set $R_i$. However, JoJoGAN overfits set $R_i$ to only a single reference style (left), thus requiring multiple generators for multiple styles. On the other hand, our Style Transformation Network maps the $\mathcal{S}$-codes to different regions of style space, resulting in a \emph{multistyle space}. Each transformed set $\hat{R}_i$ can now be overfit to a different reference style, thus allowing a single generator to learn multiple styles.}}
		\label{fig:dia2}
	\end{figure*}
	Image stylization aims to modify the \emph{style} of a given image. This is typically done by transferring the style from an exemplar style reference image to the real input image. With the advent of deep learning, various CNN-based neural style transfer algorithms are proposed to tackle the challenge of image stylization~\cite{Gatys_2016_CVPR,Zhu2017UnpairedIT,Huang2018MultimodalUI}. However, these approaches require a large amount of data consisting of both the reference style images and real images in order to train the model. In most cases, obtaining multiple examples of a particular artistic style is extremely difficult, and methods that work with as low as one example of reference style are desired.
	
	Owing to their remarkable success in synthesizing high resolution images, using a pre-trained GAN as a prior can eliminate the need of large training datasets for image stylization tasks. In recent literature, methods such as JoJoGAN~\cite{jojogan}, Mind The Gap (MTG)~\cite{zhu2022mind}, and OneshotCLIP (OSC)~\cite{Kwon2022OneShotAO} showed that it is possible to achieve desirable image stylizations by fine-tuning a pre-trained StyleGAN2 using only \emph{one} example of the reference style. For example, in the case of JoJoGAN (Fig.~\ref{fig:dia2} (left)),  it inverts the single style reference, and produces a set of similar codes using style-mixing property of StyleGAN2. During the fine-tuning step on StyleGAN2, this set is overfit to the reference image to restrict the generator to generating images of only one style. 
	
	While these methods take a promising step in improving the efficacy on one-shot image stylization task, they are limited by the fact that they require a new generator and fresh re-training for each new reference style. Storing and fine-tuning a large model like StyleGAN2 generator is not straightforward as it requires $\approx150$ MB of disk space and $> 6$ GB of GPU memory, and takes several minutes for fine-tuning for a single reference style. Many applications require to provide the users with multiple choices for the stylizations, and with existing methods, adding each new choice would raise the burden on storage and compute resources. On the other hand, large-scale StyleGAN2 models are known to have very high representation capacity, and using an entire network for producing only one stylization is surely a case of under-utilization. Moreover, the fine-tuned model becomes so specific that it is unable to generate anything else apart from the style it has fine-tuned on.
	
	To tackle this challenge, we propose a simple yet effective extension that we refer to as \textbf{MultiStyleGAN}. Our method allows for producing \emph{multiple} stylizations using a \emph{single} model that requires to be fine-tuned only \emph{once}. We depict the intuition behind our approach in Fig.~\ref{fig:dia2} (right). The key component of our method is a learnable transformation module called Style Transformation Network (STN). As described in Fig.~\ref{fig:dia2}, STN takes latent codes as input, and learns linear mappings to different regions of the latent space to produce distinct codes for each style, resulting in a \emph{multistyle space}. 
	Our lightweight transformation network ensures that the capacity of StyleGAN2 is leveraged more fully than in JoJoGAN. Thus, instead of one style per model like JoJoGAN, we are able to accommodate multiple styles while still maintaining the stylization quality (Fig.~\ref{fig:banner1}). 
	In fact, due to presence of multiple styles, our approach inherently reduces overfitting, improving the quality over existing methods. Since the model is to be fine-tuned only once, it brings down the computational requirements. Stylizing for $40$ unique styles on Nvidia A40 takes only $21$ minutes with single generator for our method as compared to total $45$ minutes, $155$ minutes, and $1400$ minutes for JoJoGAN, MTG, and OSC respectively. It also reduces the storage space requirements from $6000$ MB to $180$ MB. Further, with all styles being produced by a single generator, it allows our model to generate novel, unseen stylizations.
	
	We describe our method in Sec.~\ref{sec:approach}. In Sec.~\ref{sec:expt}, we provide comparisons of our method with three state-of-the-art one-shot stylization methods: JoJoGAN~\cite{jojogan}, Mind the Gap~\cite{zhu2022mind}, and OneshotCLIP~\cite{Kwon2022OneShotAO}. We also conducted user studies as discussed in Sec.~\ref{sec:userstudy}, which indicates that $75.00\%$, $76.11\%$, and $84.44\%$ of participants preferred our method over JoJoGAN, MTG, and OSC respectively. Further, we provide insights into several aspects of our method in Sec.~\ref{sec:analysis}.

	\begin{figure*}
		\begin{center}
			\includegraphics[width=0.97\linewidth]{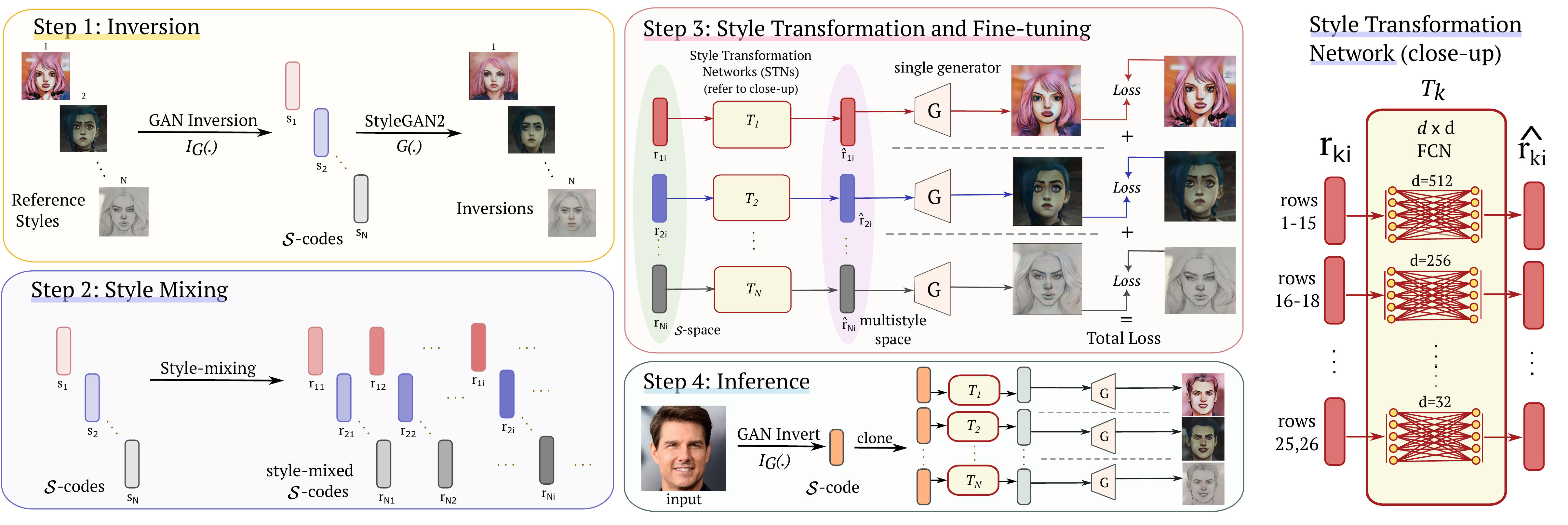}
		\end{center}
		\caption{\small{\textbf{Our Approach.} 1) Invert each style reference to obtain corresponding $\mathcal{S}$-code. 2) Apply style-mixing on the inverted codes to obtain a set $R_i$s of similar $\mathcal{S}$-codes. 3) Apply STN on set $R_i$s to transform them to distinct sets $\hat{R}_i$s in \emph{multistyle} space, and fine-tune the generator using the loss on the given sets. 4) At inference, invert the given image, and pass it through STN followed by generator to synthesize required stylizations. Since the rows of $\mathcal{S}$-code vary in their dimensionality, our STN consists of multiple fully connected networks -- one for each possible dimension ($5$ in our case) as shown in the close-up.}}
		\label{fig:dia}
	\end{figure*}

	\section{Related Work}
	\noindent \textbf{Style Transfer.} The earliest references to the style transfer are found in~\cite{efros2001image}. 
	A breakthrough in arbitrary style transfer is achieved by~\cite{gatys2016image} that leverages a CNN model to combine the style of the reference image with the content of the target image.~\cite{johnson2016perceptual} further improved the neural style transfer method by introducing perceptual losses. Since then, significant efforts are made to improve the neural style transfer methods further~ (\cite{huang2017arbitrary, WCT-NIPS-2017,park2019arbitrary}). However, in general, such methods require significant compute along with large amount of paired (or unpaired) data for the training, and are bound to fail in few-shot settings (less than $10$ examples of reference style). In contrast, our model learns to transfer multiple styles using only a single image of each.
	While the area of style transfer is explored widely~(\cite{Huang2017ArbitraryST, Li2017UniversalST, Park2019ArbitraryST}), we focus particularly on GAN-based one-shot stylization approaches that gained popularity in recent times.
	
	\noindent \textbf{One-shot Face Stylization using GANs.}
	StyleGAN models~\cite{Karras2020AnalyzingAI, Karras2020TrainingGA, Karras2019ASG} are highly successful at mimicking real data distributions such as faces. 
	Not only do they produce high quality photo-realistic images, but also provide latent spaces that are conducive to image editing applications~\cite{Shen2022InterFaceGANIT,Shen2021ClosedFormFO, Wu2021StyleSpaceAD,Tzelepis2021WarpedGANSpaceFN}.
	A pre-trained StyleGAN2 acts as a strong prior on semantic and structural detail of the image distribution~\cite{Menon2020PULSESP, Bora2017CompressedSU,Wang2021TowardsRB}. Various attempts have been made to leverage such information for one-shot/few-shot face stylization and domain adaption~\cite{Liu2021BlendGANIG, Gal2021StyleGANNADACD}.
	In general, these approaches start with a pre-trained generator of StyleGAN2, and fine-tune it using various losses and regularizations. For example, GenDA~\cite{oneshotadaption} adapts the generator to the domain of reference style by adding two lightweight classifiers, but lacks the one-to-one correspondence between source and target domain thus unable to perform image stylization. Ojha \etal~\cite{Ojha2021FewshotIG} introduces cross-domain consistency loss as a regularization in order to preserve the (dis-)similarities between the source and the target domain. While their method preserves the one-to-one correspondences and thus can be used for image stylization, it requires at least $10$ reference images, and can accommodate only one style per model.
	
	Recently, CltGAN~\cite{Wang2022CtlGANFA} proposed a few-shot technique with contrastive transfer learning strategy. OneshotCLIP~\cite{Kwon2022OneShotAO} employs CLIP \cite{Radford2021LearningTV} space consistency loss between the source and target domain generators in order to achieve one-shot adaption. Mind the gap~\cite{zhu2022mind} also leverages CLIP to determine the domain gap between source and target domain, and provide regularization accordingly to prevent overfitting. JoJoGAN~\cite{jojogan} uses style-mixing property of StyleGAN2 to obtain a paired training data for a given reference style. While providing strong results on image stylization tasks, all four of the above approaches can accommodate only one style reference at a time, and require a new model with fresh fine-tuning for each new style reference. Working with only one style also results in overfitting. Moreover, Mind the Gap~\cite{zhu2022mind} and OneshotCLIP~\cite{Kwon2022OneShotAO} also require large-scale CLIP modules for the training, resulting in increased computational burden and very high training times. On the other hand, our novel approach is compute-efficient, takes only a fraction of amount of time to train, handles multiple styles at once (up to of $120$), and reduces overfitting to produce improved results. 
	
	More recently, Pastiche Master method~\cite{pastiche} proposed DualStyleGAN -- a modified styleGAN architecture with two style paths: one for controlling the content, and one for controlling the style. While DualStyleGAN is capable of performing image stylizations, it is an intricate model designed to tackle broader set of applications. Because of that, it requires multiple non-trivial modifications in StyleGAN2 architecture, hours-long progressive training, and more importantly, a large dataset ($\approx 200$) of stylized images along with peripherals such as facial destylization module for training. Moreover, their image stylization feature works only for the styles that are already included in the training set. Contrary to that, ours is a straightforward, lightweight approach designed specifically to produce multiple one-shot stylizations at once, and can work with any set of style reference in just few minutes.

	\section{Our Approach}
	\label{sec:approach}
	\subsection{Preliminaries}
	\noindent \textbf{StyleGAN2.} StyleGAN2~\cite{Karras2020AnalyzingAI, Karras2020TrainingGA} generates diverse set of photo-realistic images by learning a mapping from random latent vectors to a high-dimensional image manifold. Unlike other GANs, it doesn't map the random vectors directly to the images, but instead first transforms the latent codes to an intermediate space known as $\mathcal{W}+$ space. Codes in $\mathcal{W}+$ space are further transformed to Style space ($\mathcal{S}$-space) which in turn is used to control image attributes. $\mathcal{S}$-space of StyleGAN2 is known to be disentangled, and offers impressive control and attribute editing capabilities~\cite{shen2020interfacegan,Tzelepis2021WarpedGANSpaceFN, Wu2021StyleSpaceAD}. Similar to JoJoGAN, our method leverages the style mixing property of $\mathcal{S}$-space to obtain a set of similar codes that can be overfit to a single reference style. Our method can also operate in other disentangled intermediate spaces offered by StyleGAN2 such as the intermediate latent space $\mathcal{W}+$.
	
	
	\noindent \textbf{JoJoGAN.} Our approach is built upon the methodology of JoJoGAN. As shown in Fig.~\ref{fig:dia2} (left), it inverts the single style reference, and produces a set of similar codes $R_i$ using style-mixing property of StyleGAN2. The key assumption is that the given style reference image can act as groundtruth stylization for all the the images obtained using style-mixing. This way, the style-mixing images along with the style reference acts as a paired dataset for stylization. Finally, the generator is fine-tuned using a combination of losses on this paired dataset. In other words, JoJoGAN fine-tunes the StyleGAN2 in a way that all the codes in set $R_i$ are mapped to the style reference. 

	\subsection{MultiStyleGAN}
	In this work, we are specifically interested in extending the approach of JoJoGAN to learn one-shot stylizations for multiple styles using a single generator. For multiple styles, fine-tuning the generator to overfit multiple sets $R_i$ to their corresponding style references is non-trivial, since each style reference may guide the generator weights in different directions. As a remedy, MultiStyleGAN obtains a new latent space called multistyle space, in which different styles lie in different regions of the space (Fig.~\ref{fig:dia2}). Such transformed sets $\hat{R}_i$ are obtained for each reference style, and can now be mapped to a corresponding reference image in order to achieve the stylizations. We describe our method in detail as follows.
	
	Let $G$ denote a StyleGAN2 generator trained on the input domain with parameters $\theta$, $I_G$ denote GAN inversion on generator $G$, $s_i$ s denote latent codes in $\mathcal{S}$-space, $u$ denote input image, and $x_k^{(ref)}$s with $k={1,2,\dots, N}$ denote $N$ style references - each representing a unique style. Our aim is to fine-tune $G$ in a way that for any input image $u$, it can produce multiple stylizations at once. For that purpose, we also introduce Stylization Transformation Networks $T_k$ with $k=1,2,...,N$. 

	Similar to JoJoGAN, we proceed in four steps (Fig.~\ref{fig:dia}), where the steps $1$ and $2$ generate a training set of paired images, and steps $3$ \& $4$ fine-tune the generator to reference styles. Unlike JoJoGAN, we proceed with multiple unique style references for steps $1$ and $2$. We also apply style transformation using STNs in step $3$ (for training), and step $4$ (for inference).
	
	\textbf{1. GAN Inversion:} In this step, all the style references are inverted into the $\mathcal{S}$-space using a suitable GAN inversion technique. We obtain $s_k^{(ref)} =S\left( I_G\left(x_k^{(ref)}\right)\right)$ where $k=\{1,2,\dots, N\}$, and $S$ represents the style mapping from $\mathcal{W}+$ space to $\mathcal{S}$-space.
	
	\textbf{2. Style Mixing}: We perform style-mixing on each of the inverted codes obtained from step 1. Some rows of each style code $s_k^{(r)}$ are mixed with randomly sampled latent codes to produce a set $R_k$ of codes $r_{ki}$ that are close to $s_k^{(ref)}$:
	\begin{align}
		r_{ki} =M \cdot s_k^{(ref)} + (1 - M) \cdot  S\left(P(z_i)\right),
		\label{eq:sty-mix}
	\end{align}  
	where $M \in \{0,1\}^{26}$ denotes a fixed mask that determines which rows of $\mathcal{S}$-space participate in style-mixing, $z_i$s are random Gaussian samples, $P$ is the mapping network, and $S$ is the style mapping from $\mathcal{W}+$ to $\mathcal{S}$-space. Note that unlike JoJoGAN, in our case, we would obtain $N$ unique sets ($R_1, \dots, R_k$), one for each reference style. 
	
	\textbf{3. Style Transformation and Fine-tuning}: Our aim is now to fine-tune the generator $G$ such that it maps all the codes in one set $R_k$ to the corresponding style reference $x_k^{(ref)}$. It isn't straightforward to achieve this since all the $R_k$ lie in the same region of the $\mathcal{S}$-space, and overfitting on one set may result in underfitting on the other. In order to separate out the vectors of the sets $R_k, k=1,\dots,N$, we transform each set $R_k$ as shown in Fig.~\ref{fig:dia2}.

	Such transformation is applied using learnable STNs $T_1, .., T_N$.  Each STN is parameterized by fully connected networks and is assigned to each unique style (Fig.~\ref{fig:dia2}). It is important to note that the $\mathcal{S}$-space code in StyleGAN2 consists of $26$ rows that vary in their dimensionality. While the first $15$ rows are $512$ dimensional, the dimensions of remaining rows decreases gradually to $256, 128, 64$ and $32$. For that reason, each STN contains $5$ fully connected networks, one for each unique dimension (see Fig.~\ref{fig:dia2}). This way, $T_k$ acts on each row of the $\mathcal{S}$ code $r_k^{(i)}$ with a single fully connected network, and linearly transforms it to produce stylized latent code $\hat{r}_k^{(i)}$:
	\begin{align}
		\hat{r}_k^{(i)} = T_k \left(r_k^{(i)}\right).
	\end{align}
	
	Thus we get $N$ sets of stylized codes in the multistyle space. As depicted in Fig.~\ref{fig:dia2}, $T_k$s are responsible for separating sets $R_k$s into new sets $\hat{R}_k$s in multistyle space. Since $\hat{R}_k$s are now distinct, each of them can be mapped to the corresponding style reference $x_k^{(ref)}$ during fine-tuning step. Note that each STN is a collection of learnable fully connected networks, and is learnt during the fine-tuning step.

	We use the following loss function in order to fine-tune the generator and to learn STNs ($T_k$s):
	\begin{align}
		\mathcal{L}_{total} =  \sum_k  L\left(G(\hat{r}_k^{(i)}) , x_k^{(ref)} \right),
	\end{align}
	where $L$ is a loss function discussed in Sec.~\ref{sec:lossfn} in detail.
	
	\textbf{4. Inference}: Once we obtain the updated generator weights $\hat{\theta}$ and the STNs $T_k$s,  stylized images can be generated as follows: First, invert the input image $u$ to get the corresponding $\mathcal{S}$-code $s^{(u)} = S\left(I_G(u)\right)$. Pass $s^{(u)}$ through STNs $T_k$s to obtain stylized codes $\hat{s}^{(u)}_k$, and finally, pass the $\hat{s}^{(u)}_k$ through fine-tuned generator to obtain stylized images $v_k$ where $k = 1,2,...,N$. Before passing it to the STNs, $s^{(u)}$ is first copied $N$ times so that each copy can be transformed using a separate transformation $T_k$. 
	Overall: 
	\begin{align}
		v_k = G_{\hat{\theta}}(T_k\left(S(I_G(u))\right)).
	\end{align}
	
	\subsection{Design of Style Transformation Network}
	We use $N$ STNs ($T_1, T_2,..., T_N$) that are independent of each other. Each of these $N$ networks are parameterized by a set of fully connected networks operating independently. Since the dimensionality of the rows of $\mathcal{S}$-code decreases gradually from $512$ to $32$ in the steps of $2$, our STN consists of $5$ fully connected networks-- one for each possible dimension of the $\mathcal{S}$-code row (Fig.~\ref{fig:dia}). Input and output dimensions of STN are same as dimension of each row of the $\mathcal{S}$-code. We initialize the weights of all the fully connected networks as identity matrix so that at the beginning of the fine-tuning, the original behavior of the generator remains preserved (See Supplementary for discussion on initialization). Further, we do not use any non-linearity in the the STN.
	
	$\mathcal{W}+$ space of StyleGAN offers similar disentanglement properties as the $\mathcal{S}-$space, thus making it possible for our model to operate in $\mathcal{W}+$ space as well. In that case, the STN requires only a single fully connected network, since all the rows in the $\mathcal{W}+$ code are of same dimensions ($512$). 
	
	\subsection{Loss Function}
	\label{sec:lossfn}
	To keep our method simple, we directly follow the loss function design of JoJoGAN for our model. The loss function consists of $l-$1 loss computed over the discriminator features i.e. discriminator perceptual loss:
	\begin{align}
		L_D = \sum_k \| D(G(\hat{r}_k^{(i)})) -  D( x_k^{(ref)}) \|_1
	\end{align}
	As discussed in~\cite{jojogan}, identity loss~\cite{Deng2019ArcFaceAA} and contextual loss~\cite{Mechrez2018TheCL} can be added optionally for improved performance. In our experiments, we choose to use discriminator perceptual loss and contextual loss with weight factor of $0.005$.

	\section{Experiments}
	\label{sec:expt}
	
	\begin{figure*}
		\begin{center}
			\includegraphics[width=0.95\linewidth]{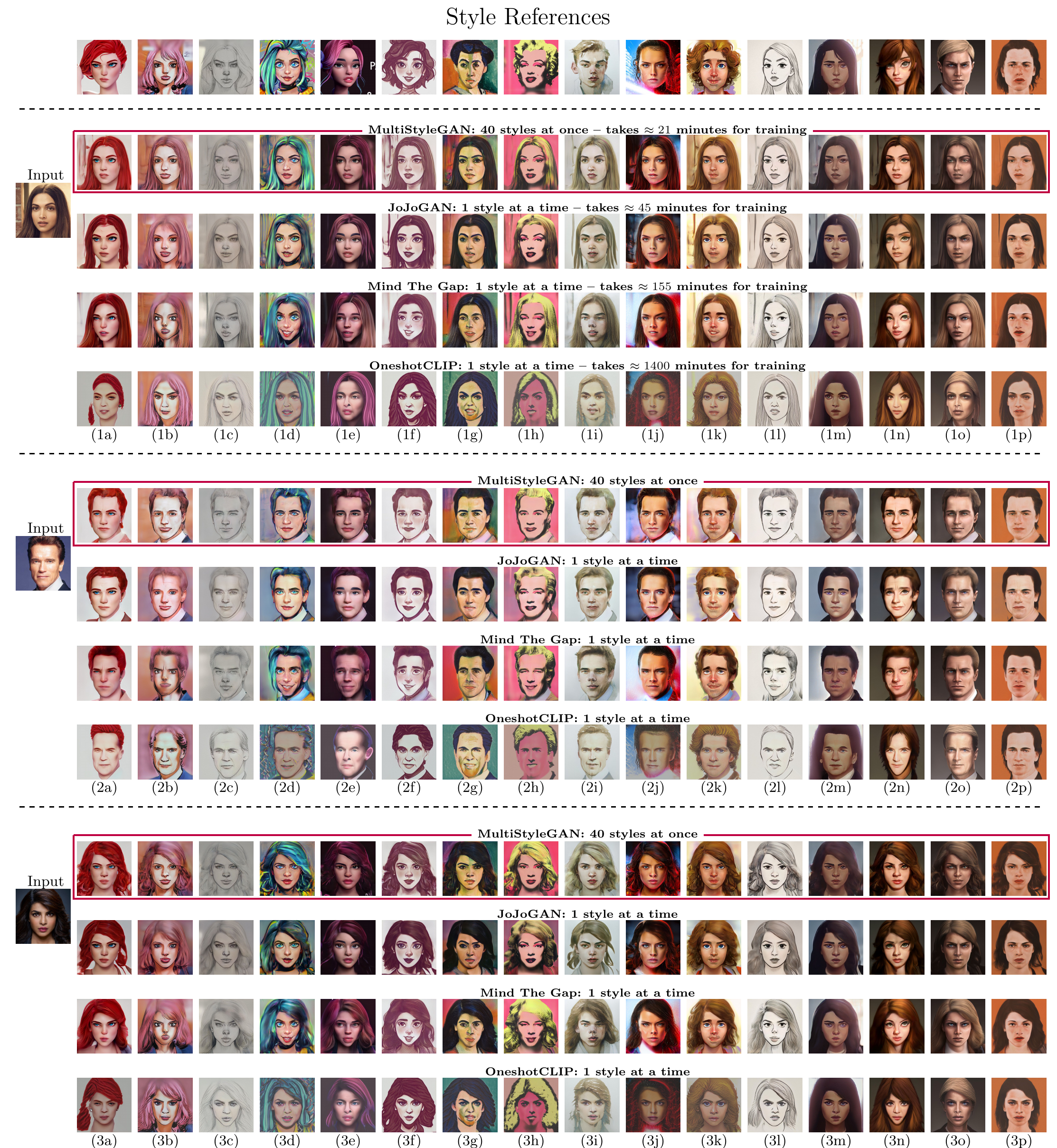}
		\end{center}
		\caption{\small{\textbf{Comparisons.} 
				Apart from producing $40$ stylizations at once, MultiStyleGAN brings noticeable improvements over current state-of-the-art methods. MultiStyleGAN prevents overfitting and preserves identity \& expressions of the inputs. JoJoGAN overfits on reference identity/expressions (see cols. 1i, 1k, 3f); MTG distorts the shape of face (see cols. 1b), chin (see cols. 1e, 1d), and nose (see col. 1l); OSC produces incomplete faces (see cols. 1a, 1o) and distorts the expressions (see cols. 1l, 1i). Full set of results are provided in Supplementary.}}

\label{fig:compare}
\end{figure*}

In this section, we compare with two existing methods for one-shot image stylization. First, we compare with JoJoGAN since our approach builds upon it, and as they show significant improvements over previous state-of-the-art methods. Second, we compare with the most recently published method Mind the Gap (MTG). We provide comparisons with OneshotCLIP (OSC)~\cite{Kwon2022OneShotAO} as well. 
We also discuss various aspects of our approach such as the effect of increasing the number of reference styles, and generating novel stylizations.
\subsection{Implementation Details} 
\label{sec:impl}
For all the experiments, we start with a pre-trained StyleGAN2 trained on FFHQ dataset at $1024 \times 1024$ resolution.

Similar to JoJoGAN, there are several design choices to make in MultiStyleGAN methodology as well. For example, choice of GAN inversion method $I_G$, style-mixing mask $M$, loss function, and the latent space to operate in. Since our setup is based on JoJoGAN, we rely on the extensive discussions provided in the JoJoGAN paper to determine our setup. In fact, we keep our choices simple and uniform across all experiments with both MultiStyleGAN and JoJoGAN: we use e4e~\cite{tov2021designing} for inversion and operate in $\mathcal{S}$-space; out of $26$ total rows in $\mathcal{S}$-codes, we select $14$ rows for applying style-mixing with $M=\{0~\text{if}~i<12~\text{else}~1\}$. We use the combination of discriminator loss and contextual loss (with weight $0.005$); and use $500$ iterations of Adam optimizer at learning rate of $0.002$ for the generator. We noticed that setting up a lower learning rate for the STN helps stabilizing the fine-tuning, thus we use learning rate of $1e-5$ for the parameters of STN. Further, we choose to initialize the STNs using identity matrix.

Note that in principle, all the variants of JoJoGAN described in~\cite{jojogan} still remains applicable for MultiStyleGAN. For example, even in MultiStyleGAN, aspects of styles that are being transferred can be controlled by choosing appropriate mask $M$, or extent of stylization can be controlled by changing the GAN inversion method as described in~\cite{jojogan}. 
We defer the discussion of such variations for MultiStyleGAN to Supplementary. 

For Mind the Gap and OneshotCLIP, we use the default settings from their official implementations. 
\subsection{Results}

\begin{figure*}[t]
	\centering
	\includegraphics[width=.92\linewidth, trim=0 63.7em 0 0, clip]{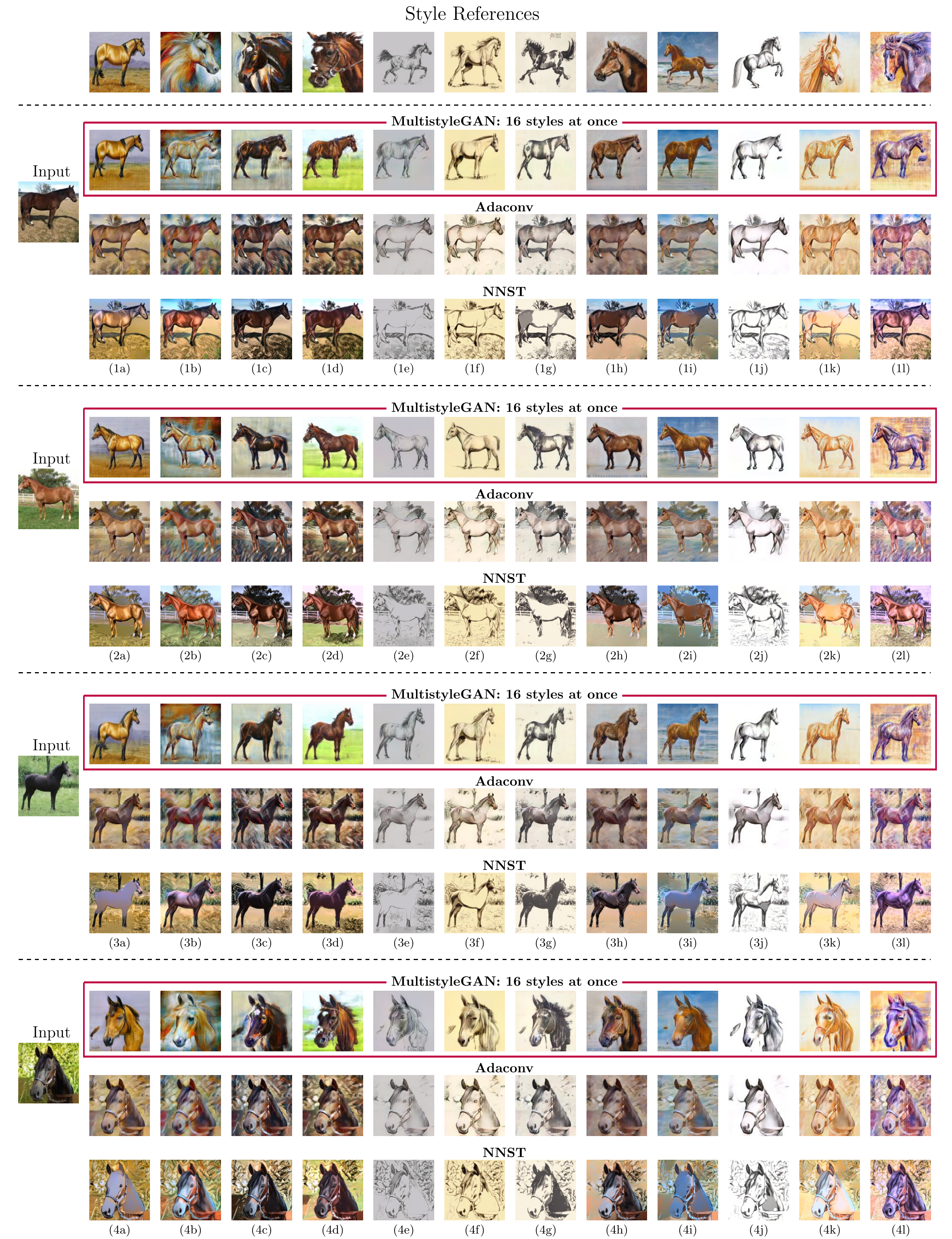}
	\captionof{figure}{\small \textbf{Comparisons on  Horses. } MultiStyleGAN can successfully stylize non-face inputs such as horses outperforming other generic style transfer methods. The above results are obtained with MultistyleGAN  in $\mathcal{S}$-space with $N=16$.}
	\label{fig:horse}
\end{figure*}

\begin{figure*}[!h]
	\centering
	\includegraphics[width=.98\linewidth, trim=0 63.7em 0 0, clip]{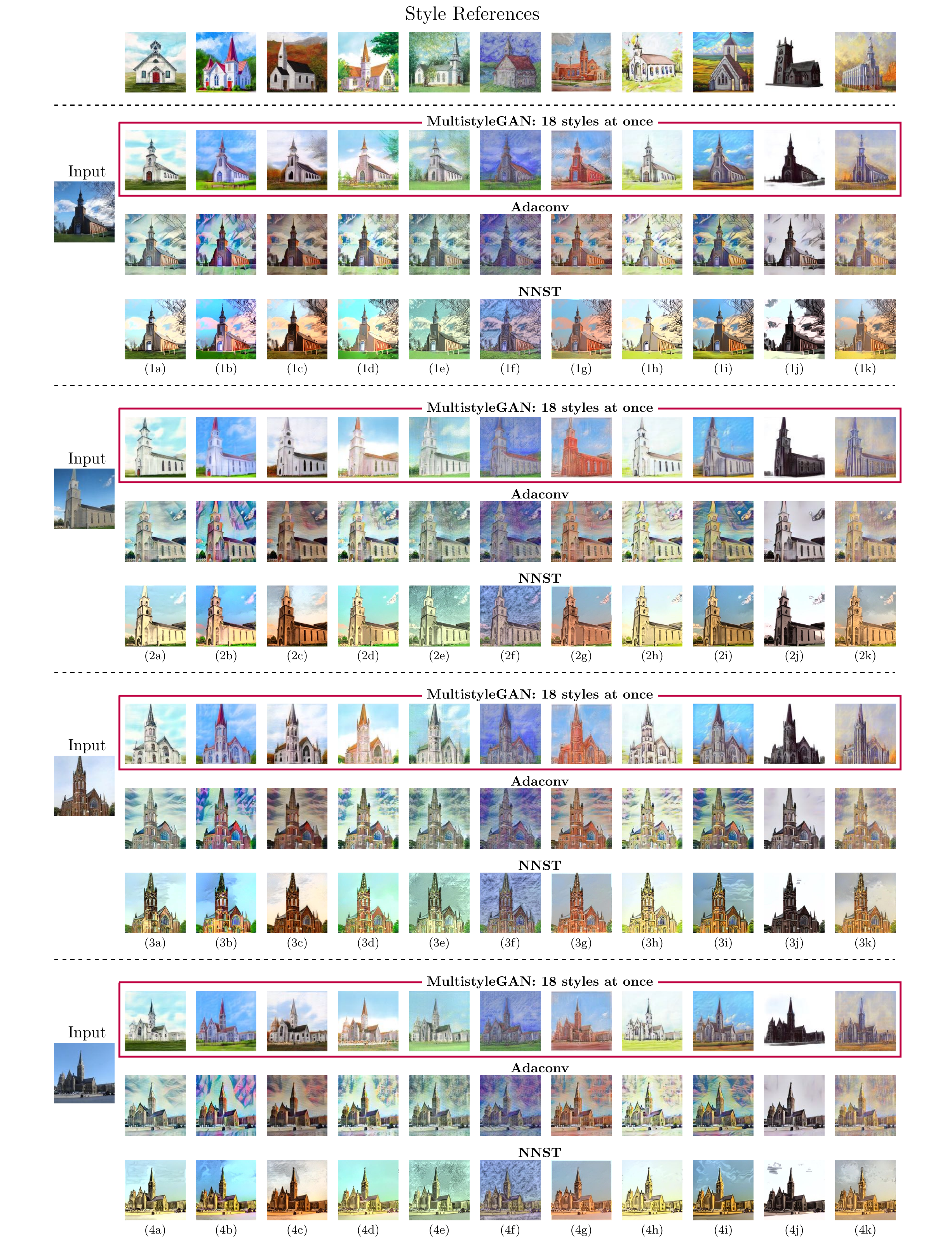}
	\captionof{figure}{\small \textbf{Comparisons on Churches. } MultiStyleGAN can successfully stylize non-face inputs such as churches outperforming other generic style transfer methods. The above results are obtained with MultistyleGAN model in $\mathcal{S}$-space with $N=18$.}
	\label{fig:church}
\end{figure*}

\begin{figure*}[t]
	\centering
	\begin{minipage}[t]{.35\textwidth}
		\centering
		
		\captionof{table}{\scriptsize Comparison of training time for $N$ styles}
		\label{tab:runningtime}
		\vspace{-1em}
		\begin{center}
			{\scalebox{0.7}{
					\begin{tabular}{lccc}
						\toprule
						& \multicolumn{3}{c}{Total Training Time (in hh:mm:ss)} \\
						\cmidrule(lr){2-4}
						Method & $N=40$ & $N=80$ & $N=120$ \\
						\midrule
						JoJoGAN~\cite{jojogan}&  00:45:20 & 01:30:40 & 02:16:00 \\
						Mind the Gap~\cite{zhu2022mind}& 02:34:36 & 05:09:20&  07:43:56   \\	
						OneshotCLIP~\cite{Kwon2022OneShotAO}& 24:01:06 & 48:02:12  &  96:03:18 \\		
						MultiStyleGAN (ours)  &  \textbf{00:21:32} & \textbf{00:41:58} & \textbf{00:57:02}  \\
						\bottomrule
			\end{tabular}}}
		\end{center}
	\end{minipage}%
	\hfill
	\begin{minipage}[t]{.32\textwidth}
		\captionof{table}{\scriptsize User Study Results}
		\label{tab:userstudy}
		\begin{center}
			
			{\scalebox{0.7}{\centering
					\setlength{\tabcolsep}{2pt}
					\begin{tabular}{lccc}
						\toprule
						\multicolumn{4}{c}{$\%$ Preference for MultiStyleGAN over:} \\
						\cmidrule(lr){1-4}
						\textbf{on Faces} 	& JoJoGAN & MTG & OSC  \\
						\midrule
						in $\mathcal{W}+$space & $72.8\%$ & $71.2\%$  & $73.8\%$ \\
						in $\mathcal{S}$-space &  $75.0\%$ & $76.1\%$ & $84.4\%$  \\
						\bottomrule
						\addlinespace
						\textbf{on Non-faces} &  & AdaConv & NNST  \\
						\cmidrule{1-4}
						in $\mathcal{S}$-space &  & $78.2\%$ & $72.3\%$   \\
						\bottomrule
			\end{tabular}}}
		\end{center}
	\end{minipage}%
	\hfill
	\begin{minipage}[t]{.32\textwidth}
		\centering
		
		\captionof{table}{\scriptsize SIFID~\cite{rottshaham2019singan} Scores Comparison}
		\label{tab:sifid}
		\begin{center}
			\scalebox{0.7}{
				\begin{tabular}{lc}
					\toprule
					Method & SIFID$\downarrow$ \\
					\midrule
					JoJoGAN~\cite{jojogan}& 0.966  \\
					Mind the Gap~\cite{zhu2022mind}&  1.007 \\	
					MultiStyleGAN; $\mathcal{W}+$space; $N=40$ &  0.892 \\		
					MultiStyleGAN; $\mathcal{S}$-space; with $N=40$ & \textbf{0.813}\\
					\cmidrule(lr){1-2}
					MultiStyleGAN; $\mathcal{S}$-space; $N=80$ & 0.905 \\
					MultiStyleGAN; $\mathcal{S}$-space; $N=120$ & 1.001\\
					\bottomrule
			\end{tabular}}
		\end{center}
	\end{minipage}
\end{figure*}

An ideal image stylization algorithm should produce outputs with the best depiction of reference style, while still preserving the expressions and identity of the input image. As shown in Fig.~\ref{fig:banner1}, MultiStyleGAN performs well on both these criteria on diverse set of style references and input images. To demonstrate its capability of accommodating diverse styles in a single model, we include variety of Face styles such as cartoon, sketch, caricature, pastel and watercolor painting. This makes the results in Fig.~\ref{fig:banner1} and Fig.~\ref{fig:compare} quite representative of the efficacy of our model, consistent with the additional results provided in Supplementary. Further, we also show results on non-face images such as churches, and horses in Fig.~\ref{fig:church} and Fig.~\ref{fig:horse}. 

\noindent \textbf{Qualitative comparisons.}
In Fig.~\ref{fig:compare}, we compare MultiStyleGAN with JoJoGAN~\cite{jojogan}, MTG~\cite{zhu2022mind} and OSC~\cite{Kwon2022OneShotAO}. Our method works with multiple styles at once without compromising the quality of the outputs. In fact, presence of multiple styles prevents overfitting -- a prevalent issue in JoJoGAN. MTG and OSC distort the facial shapes too much (chin, nose etc.), while our method is able to generate consistent shapes due to simple pixel-level losses used for fine-tuning. Qualitatively, we can see that MultiStyleGAN preserves the identity and expressions of input images more consistently than JoJoGAN, MTG and OSC. For example, for the col. (1b) in Fig.~\ref{fig:compare}, JoJoGAN, MTG and OSC distort the facial shape, hair, and expression of the input significantly, while MultiStyleGAN preserves these features in a faithful manner. Note that the MultiStyleGAN results in Fig.~\ref{fig:compare} are obtained using a model trained on $40$ styles at once using $\mathcal{S}$-space variant. While we included $16$ of those here, the remaining are provided in Supplementary.

We also provide qualitative comparisons on non-faces (horses and churches) in Fig.~\ref{fig:horse} and Fig.~\ref{fig:church}. Here, we compare our method with existing generic style transfer approaches AdaConv~\cite{Chandran2021AdaptiveCF} and NNST~\cite{Kolkin2022NeuralNS}. Results indicate that AdaConv and NNST fail to reproduce the correct colorization in their outputs and lack semantic understanding of the scene, while our method generates faithful stylizations while preserving the semantic structure of the input. We provide extensive set of results on non-face images in the supplementary.

\noindent \textbf{Training time and storage.} For JoJoGAN, MTG, and OSC, we need to fine-tune for $40$ styles individually, while in case of MultiStyleGAN, all the $40$ styles in Fig.~\ref{fig:compare} required only one model - bringing down the computational requirements drastically. We present the comparisons of total training times for all $4$ methods in Tab.~\ref{tab:runningtime} keeping number of fine-tuning steps the same ($500$) for all.  For MultiStyleGAN, total training time doesn't vary linearly with $N$. For the case with $40$ styles (Fig.~\ref{fig:compare}), it takes $\approx 22$ minutes on Nvidia A40, improving $2.2\times$ over JoJoGAN, $8\times$ over MTG, and $60\times$ over OSC. In terms of storage, each trained generator takes $\approx 150$ MB of disk space, while the lightweight STNs need negligible storage ($\approx 30$ MB). Thus, we get approximately $N\times$ improvement in storage. 

\begin{figure*}
\centering
\begin{minipage}[t]{.36\textwidth}
	\centering
	\includegraphics[width=.98\linewidth]{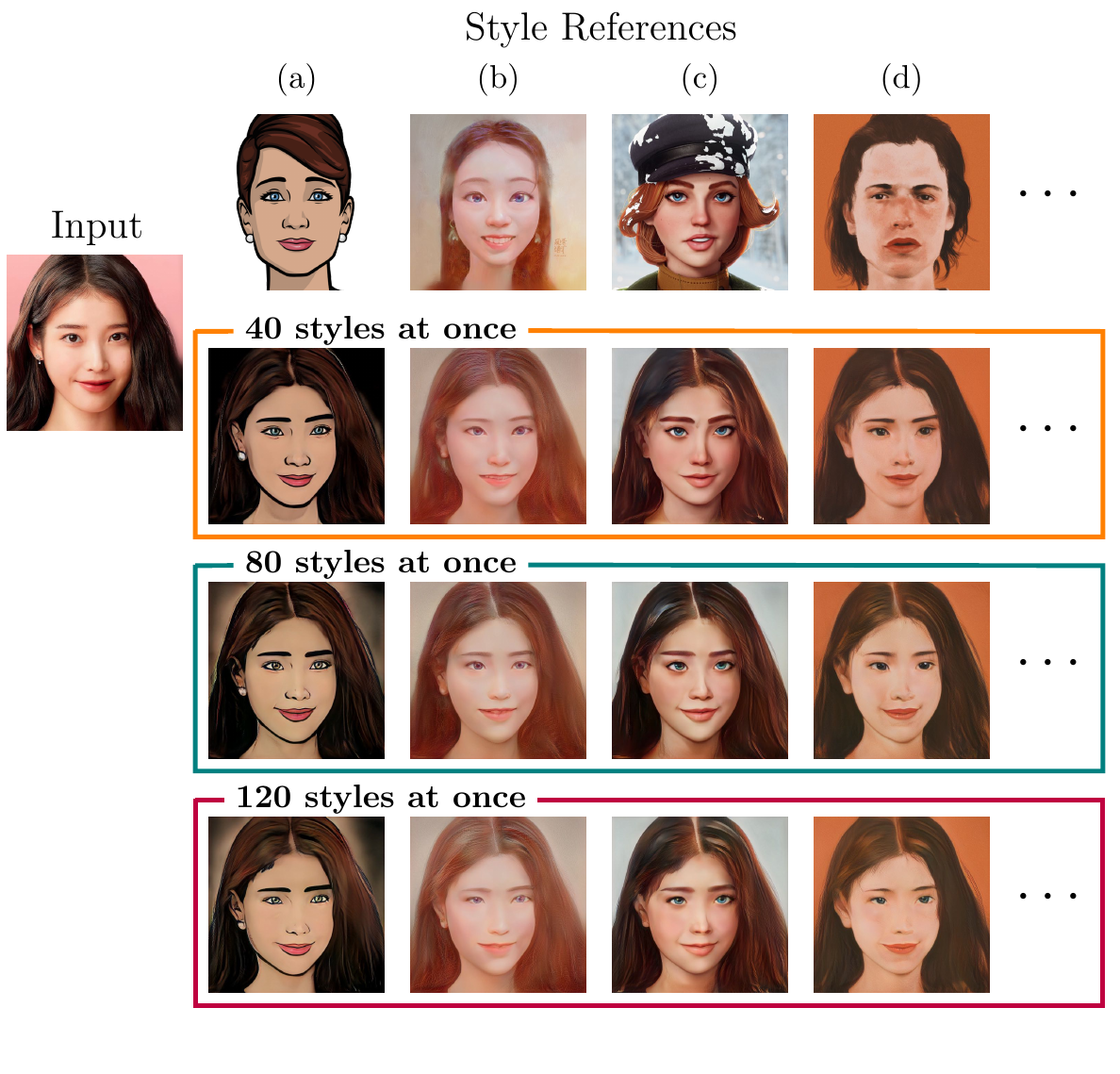}
	\captionof{figure}{\small{\textbf{Increasing total number of styles.} Barring some minor variations such as hair color in (a) and skin color in (c), quality of stylizations remain consistent for MultiStyleGAN as total number of styles $N$ are increased.}}
	\label{fig:result3}
\end{minipage}%
\hfill
\begin{minipage}[t]{.32\textwidth}
	\centering
	\includegraphics[width=.92\linewidth]{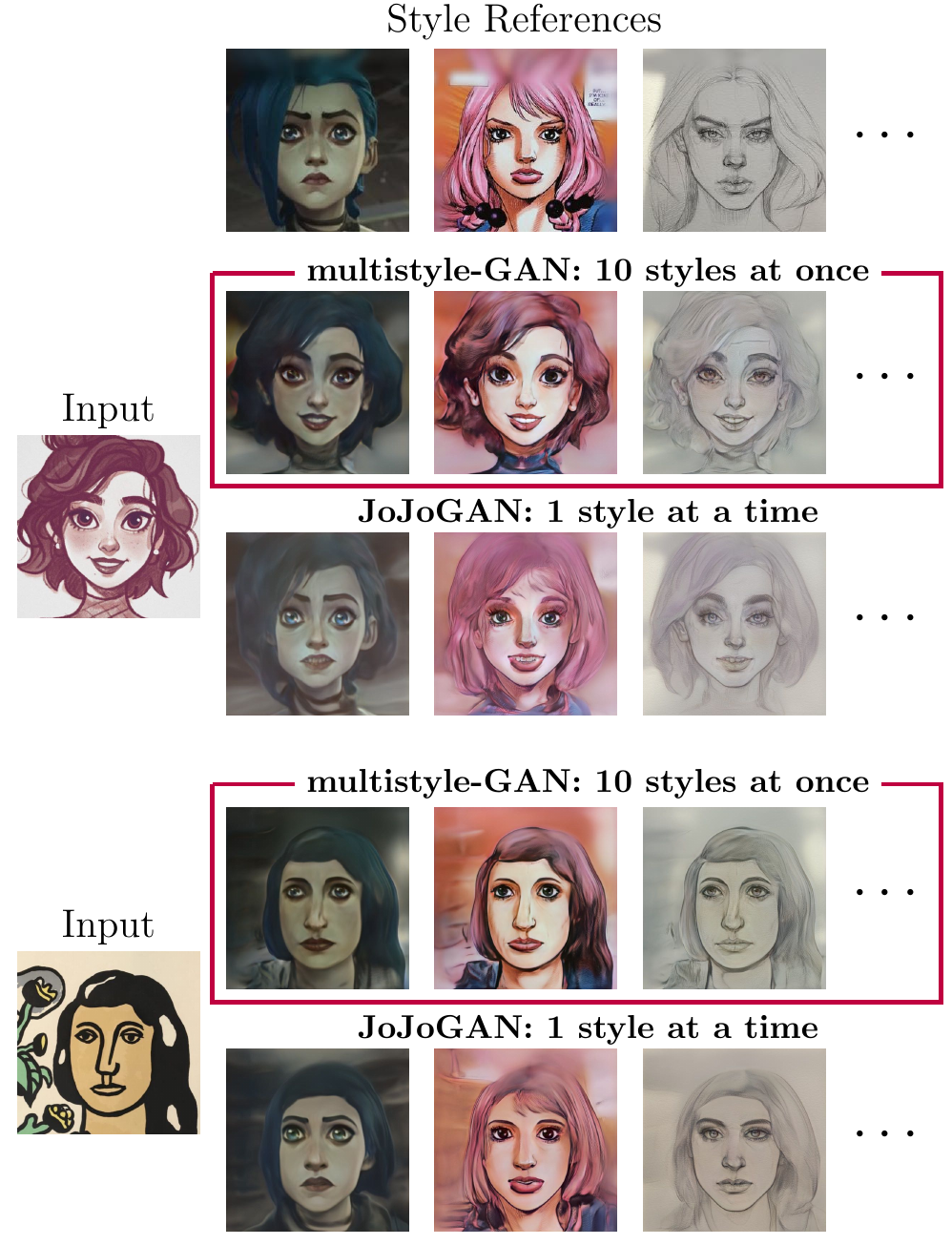}
	\captionof{figure}{\small{\textbf{Re-stylizing unseen style images.}} Our model can stylize unseen style images to the reference styles while preserving the identity. JoJoGAN fails for such out-of-domain examples.
		}
		\label{fig:cross}
	\end{minipage}%
\hfill
\begin{minipage}[t]{.27\textwidth}
\centering
\includegraphics[width=0.87\linewidth]{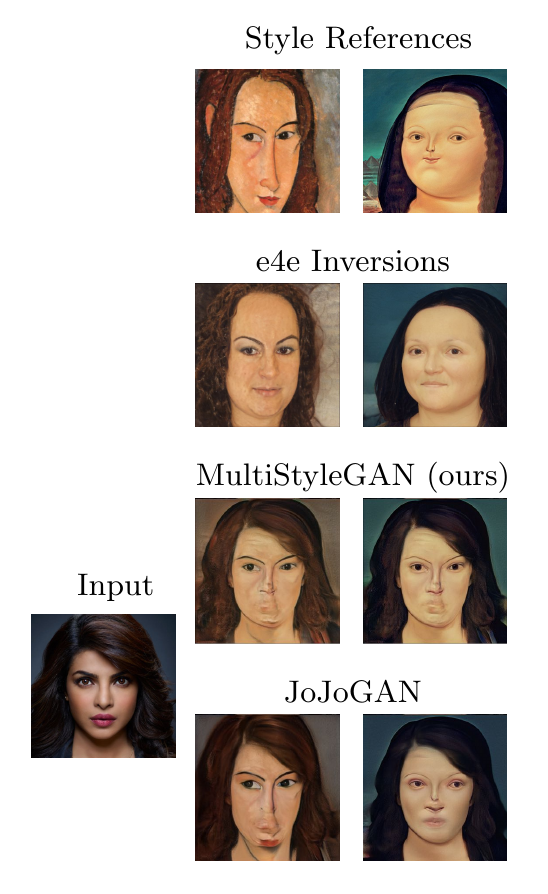}  \\
\captionof{figure}{\small \textbf{Failure cases.} Our method fails when style references have distinctively different facial shapes. This can be attributed to poor quality of GAN inversions.
} 
\label{fig:failure}
\end{minipage}
\end{figure*}

	\label{sec:results}

	\noindent \textbf{User studies.}
	\label{sec:userstudy}
	Because stylization is inherently subjective, the commonly agreed upon method for evaluating the stylization efficacy is to conduct a human evaluation study. Therefore, we conduct user studies to compare our method against JoJoGAN, MTG, and OSC. In our study, each participant is shown an input face and a reference style along with outputs of two methods being compared, in a random order, and asked which output best depicts the reference style while preserving the identity and expressions of the input. For each unique pair of input and reference, we provided two options: one generated using MultiStyleGAN, and the other generated using the comparison method. 
	
	We conducted separate user studies for comparison with JoJoGAN, MTG, and OSC.
	For each study, we received $180$ responses from $30$ users where each responded to $6$ questions. Both the reference styles and input images used in the study are chosen at random, and were kept fixed for all participants. Results indicate that $\mathbf{75.0\%}$ participants preferred our method over JoJoGAN, $\mathbf{76.1\%}$ participants preferred it over MTG, and $\mathbf{84.4\%}$ participants preferred it over OSC. This can be attributed to issues of overfitting in JoJoGAN and facial shape distortions in MTG, both of which are mitigated in our method. We repeated the user studies by replacing the $\mathcal{S}$-space variant of our model with $\mathcal{W}+$space variant, and observed similar performance improvements as shown in Tab.~\ref{tab:userstudy}. On non-face images, our method is preferred by $\mathbf{78.2\%}$ and $\mathbf{72.3\%}$ of the participants compared to AdaConv~\cite{Chandran2021AdaptiveCF} and NNST~\cite{Kolkin2022NeuralNS} respectively.
	
	\noindent \textbf{Quantitative Evaluation.} In many GAN-related applications, FID score~\cite{Heusel2017GANsTB} measured by comparing population statistics is presented as a quantitative metric, however, in the case of one-shot stylization, FID evaluation is inadequate since it 
	requires a reasonably large dataset of style images that are not available in one-shot cases. Moreover, computing FID on small datasets incur significant errors~\cite{Chong2020EffectivelyUF}. Therefore, for quantitative evaluation, we use Single Image FID (SIFID) metric~\cite{rottshaham2019singan} that is designed specifically to measure similarities in output and style references for cases where only one reference sample is available. In our case, we select $12$ styles at random for evaluation, and for each style we calculate SIFID on $15$ randomly selected input images. Average SIFID scores across all the styles are shown in Tab.~\ref{tab:sifid}, which clearly indicates the superior performance of MultiStyleGAN in both $\mathcal{W}+$ and $\mathcal{S}$-space.

	\subsection{Detailed Analysis}
	\label{sec:analysis}
	\noindent \textbf{Effect of increasing number of reference styles.}
	To analyse the trade-off between the number of styles ($N$) \textit{vs.} the quality of stylizations, 
	in Fig.~\ref{fig:result3} we provide comparisons for the MultiStyleGANs that were fine-tuned with $40, 80$ and $120$ unique styles respectively.  We can observe some variations in the stylistic features of these results: inconsistency in the hair style for (a), and slight variation in the intensity of color for (b).  Barring such minor artifacts, the quality of stylization still remains impressive for higher number of styles owing to the large capacity of StyleGAN2. More results are included in Supplementary. We include the variation of SIFID scores as we increase the number of styles for the $\mathcal{S}$-space variant of our model in Tab.~\ref{tab:sifid}. It is evident that our model maintains the quality of stylizations. 
	
	These results tell us that it should be possible to extend the method beyond $120$ styles given the large capacity of StyleGAN2. However, in practice, the number is limited by the compute resource constraints, as adding each new style will increase the compute requirement for the forward pass by a small amount. In our case, we could accommodate as high as $120$ styles at once on a single Nvidia A40 GPU with $48$ GB of RAM. More efficient implementations can potentially increase the feasible value of $N$.
	
		\begin{figure*}
		\begin{center}
			\setlength{\tabcolsep}{12pt} 
			\begin{tabular}{cc}
				\includegraphics[width=0.46\linewidth]{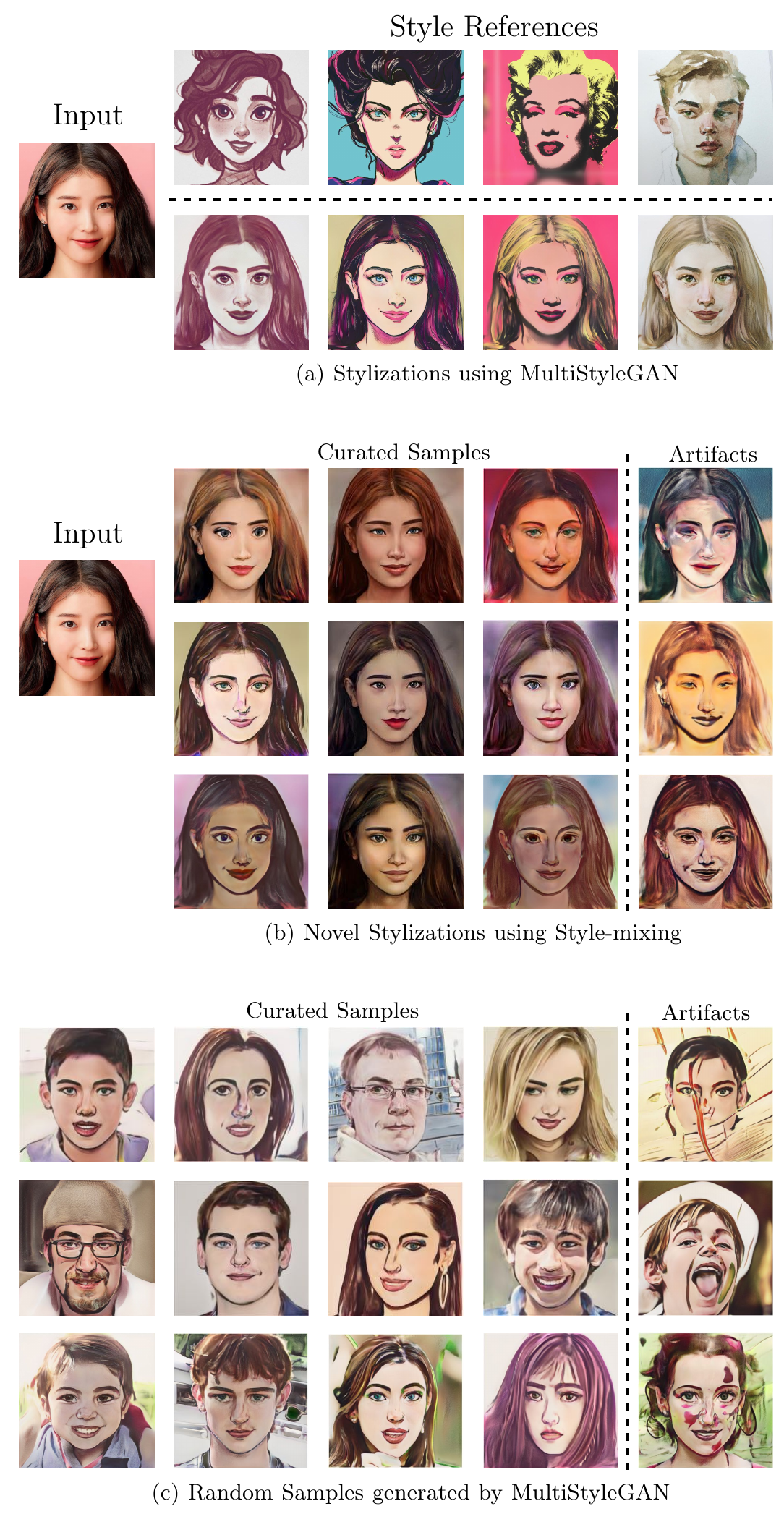} &
				\includegraphics[width=0.46\linewidth]{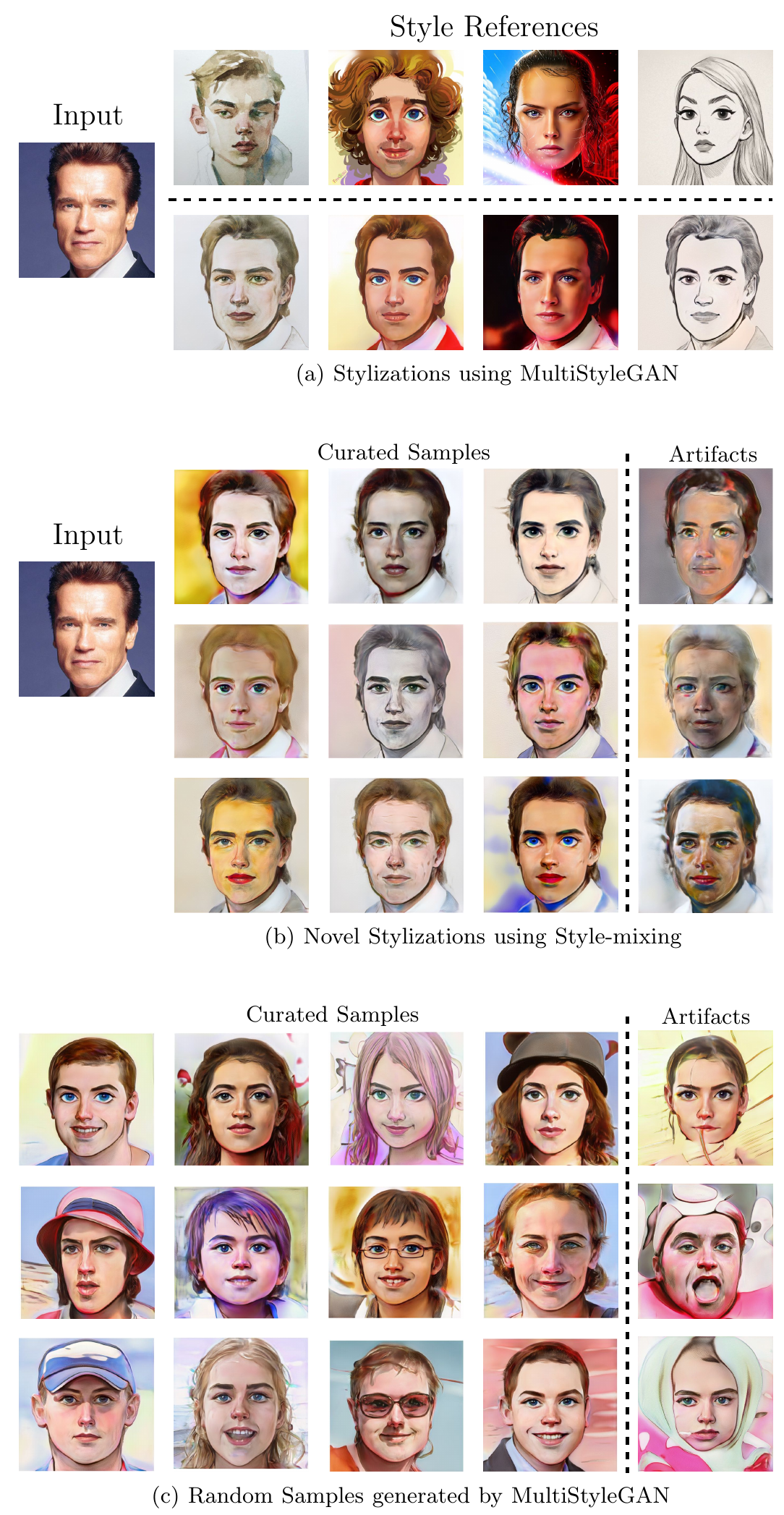}
			\end{tabular}
		\end{center}
		\caption{\small{ \textbf{Generating novel styles.} Since our model is trained using multiple styles, we can leverage the smoothness property of StyleGAN2 manifold to traverse in multistyle space and obtain new, unseen stylizations. In the above examples, we train a MultiStyleGAN model on $4$ style references and generate corresponding stylizations in (a). We then perform style mixing with random codes in multistyle space to obtain some novel stylizations as shown in (b). Further, we generate random samples from multistyle space that result in diverse set of images with distinct styles (c). Random style mixing and random generation may yield samples with artifacts as shown in the right column for both (b) and (c).}}
		\label{fig:blend}
	\end{figure*}

	\noindent\textbf{Re-stylizing the reference styles.}
	\label{sec:swap}
	One-shot stylization methods are trained to stylize real input images, thus it remains challenging to stylize out-of-domain unrealistic inputs. We consider a case of out-of-domain input where the input image itself is a stylized image of a style that hasn't been seen by the generator. In this case, we attempt to \emph{re-stylize} such style input to a new style with MultiStyleGAN. Our approach remains the same: Just like any other arbitrary input, we perform GAN inversion on the input, transforms it to multistyle space, and pass it through MultiStyleGAN to get the outputs. As depicted in Fig.~\ref{fig:cross}, even for such out-of-domain examples MultiStyleGAN preserves the identity of the input to produce faithful stylizations, while JoJoGAN yield poor results due to overfitting. In case of MultiStyleGAN, the STNs bring the out-of-domain codes in style space to in-domain codes in multistyle space, helping generator to synthesize strong outputs. Other methods do not employ STNs due to which their generators fail to synthesize high quality images from out-of-domain latent codes. Note that the input images were not used during fine-tuning of the MultisStyleGAN used to generate results in Fig.~\ref{fig:cross}. More results are provided in Fig.~\ref{fig:cross_s} and Fig.~\ref{fig:cross_matrix}.
	
	\noindent\textbf{Generating novel styles.} 
	StyleGAN2 is known for learning smooth image manifolds that can be traversed to vary image attributes of interest. Thus, one important advantage of being able to learn multiple stylizations using a single StyleGAN2 is the possibility of exploiting the network's knowledge of various stylizations to create new, unseen stylizations. We do so by applying the style-mixing property of StyleGAN2 in multistyle space of a fine-tuned MultiStyleGAN. To depict such zero-shot behavior, we train a MultiStyleGAN on $4$ styles and obtain stylizations of the input as usual (Fig.~\ref{fig:blend} (a)). Further, we perform style-mixing on the multistyle codes of the input image in a way that the style of the image varies while keeping the identity intact (see Fig.~\ref{fig:blend} (b)). This can be done by applying style-mixing only on the rows of the codes that corresponds to higher resolution layers of the StyleGAN2. It is evident that the new stylizations generated in this manner are significantly different than the ones used as a reference, thus, giving us \emph{induced zero-shot} stylizations. It should be noted since JoJoGAN, MTG, and OSC are trained to generate images of only one style, such behavior cannot be obtained using their methods.%

	Further, we sample random codes from the multistyle space and display images synthesized from them in Fig.~\ref{fig:blend} (c). Such random generation gives a more diverse set of styles, and while some elements of these styles can be traced back to the original style references, the way in which they are blended together can be unpredictable. Since both the random style mixing and random image generation are heuristic procedures, they may produce some images with artifacts. We display such ``bad'' samples in the right columns of Fig.~\ref{fig:blend} (b, c). In our experiments, we observe that the percentage of such images with artifacts are around $25$ to $30\%$ for random style mixing, and are around $15$ to $20\%$ for random generation.

	\noindent \textbf{Failure cases.} 
	\label{sec:limit}
	While our results are impressive, there remains a few limitations to be investigated. We depict some limitations in Fig.~\ref{fig:failure}: the stylization becomes difficult if the reference styles have distinctively different shapes of the faces. This can be attributed to poor quality of inversions produced by e4e~\cite{tov2021designing}. Another potential reason is that the rows of the $\mathcal{S}$-code that controls the shapes of the faces are not included in the style-mixing mask $M$ used in Eq.~\ref{eq:sty-mix}. Therefore, network cannot learn how to apply such heavy changes in the shape of the face. Tuning the mask $M$ can improve the results by some margin, but at the cost of distorting the identity of input image. Even though JoJoGAN performs slightly better owing to its overfitting behavior, this remains a common limitation of the GAN-based one-shot stylization methods. We include additional failure cases (on non-face images) in the Supplementary. 
	\section{Conclusion and Future Work}
	In this work, we propose a novel MultiStyleGAN approach for learning multiple one-shot image stylizations using only a single pre-trained generator. Our method requires only a single image of each reference style, and can accommodate as many as $120$ styles in a single fine-tuning. We compare our method with state-of-the-art one-shot image stylization methods, and show that our approach brings improvements not only along the axis of quality of the stylizations, but also along the axis of speed and storage efficiency. We also discuss novel use cases of our method, such as generating novel stylizations, and re-stylizing the reference styles. As a future work, we plan to improve on the limitations discussed in Sec.~\ref{sec:limit}.
	
	\section{Acknowledgments}
	We thank Mohit Goyal, Min-Jin Chong, and Aiyu Cui for helpful discussions and suggestions.

	{\small
		\bibliographystyle{ieee_fullname}
		\bibliography{egbib}
	}

	\clearpage
		\section*{Supplementary Material}

\begin{figure}[h!]
	\begin{center}
		\includegraphics[width=0.55\linewidth]{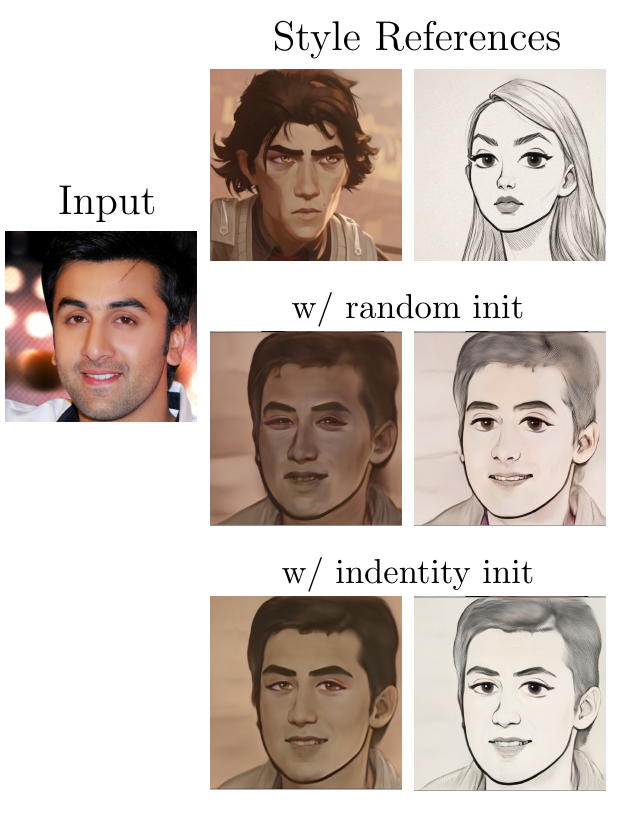} 

	\end{center}
	\caption{\small{\textbf{Initialization of STN.} Using identity matrix as init for the fully connected networks in STNs preserves the default behavior of pre-trained generator, thus the model converges faster and produce better results as compared to standard normal init.}}
	\label{fig:init}
\end{figure}

		\noindent \textbf{A.1 Initialization of STNs.} 
	\label{sec:init}
	Since we use a pre-trained generator with completely untrained STNs, the choice of initialization method for STNs becomes important. If initialized incorrectly, STNs may take a large number of iterations to converge. We obtained the best results by making sure that for completely untrained STNs, the default behavior of StyleGAN2 is preserved, \textit{i.e.}, STN weights are initialized with the identity matrix. As shown in Fig.~\ref{fig:init}, this helps our model achieve better results as compared to standard random initialization. 
	
\begin{figure*}[t!]
\begin{center}
	\includegraphics[width=0.95\linewidth]{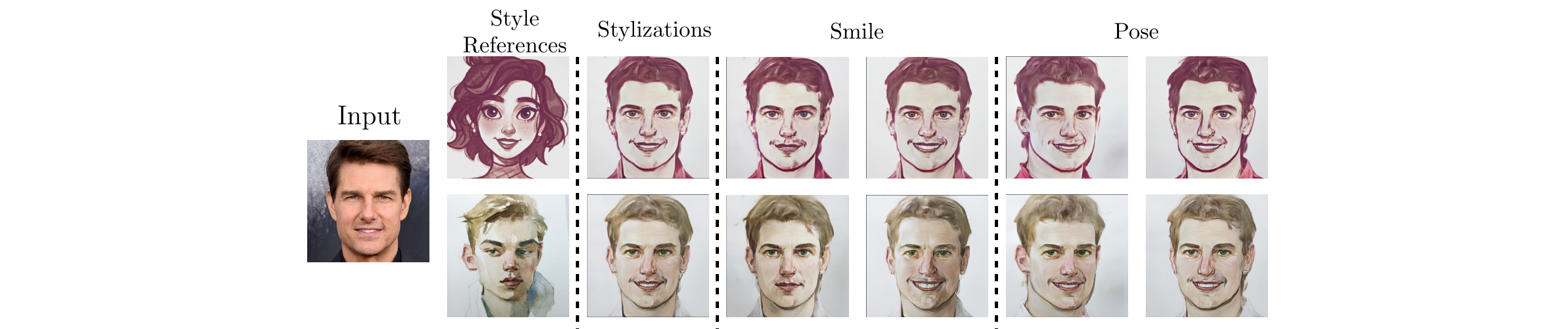}
\end{center}
\caption{\small{\textbf{Editing Stylized images using SeFa. }} MultiStyleGAN preserves the editing properties of StyleGAN2, thus allowing us to edit various attributes of the stylized image using off-the-shelf image editing techniques such as SeFa~\cite{shen2021closed} or InterfaceGAN~\cite{shen2020interfacegan}. Here, we show edits in smile and pose obtained using SeFa on MultiStyleGAN model trained on $4$ styles. }
\label{fig:edit}
\end{figure*}

	\noindent\textbf{A.2 Editing Stylized Images.}
\label{sec:sup_editing}
MultiStyleGAN preserves the smoothness of the underlying StyleGAN2 manifold, thus allowing us to edit specific attributes of stylized images using any off-the-shelf GAN image editing method such as SeFa~\cite{shen2021closed} or InterfaceGAN~\cite{shen2020interfacegan}. We provide results of image editing obtained by applying SeFa on a MultiStyleGAN model trained using $4$ styles. As shown in Fig.~\ref{fig:edit}, our model is able to produce smooth editing for both smile and pose attributes.

	\noindent \textbf{A.3 Variations of MultiStyleGAN.} 
	Similar to JoJoGAN, there are several design choices to make in MultiStyleGAN methodology as well. For example, choice of GAN inversion method $I_G$, style-mixing mask $M$, loss function, and the latent space to operate in. Each of these design choices control various aspects of stylization, \textit{e.g.} aspects of styles that are being transferred can be controlled by choosing appropriate mask $M$, or extent of stylization can be controlled by changing the GAN inversion method. While these variations are described in detail in JoJoGAN manuscript, we present experimental results for two such cases. 
	
\begin{figure}[h!]
\begin{center}
	\includegraphics[width=0.50\linewidth]{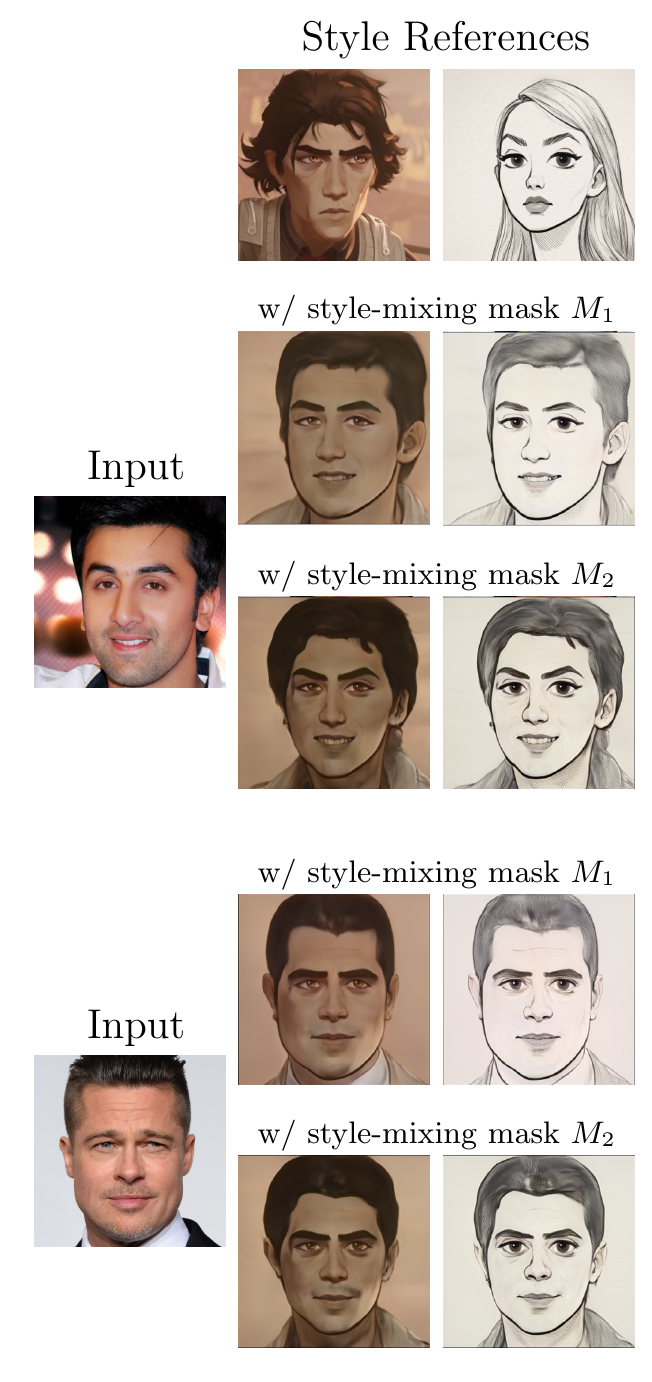}
\end{center}
\caption{\small{\textbf{Varying the style-mixing mask $\mathbf{M}$. }} Just like JoJoGAN, varying the style-mixing mask $M$ gives us control over what aspects of styles are transferred from the reference to the input. In this example, mask $M_1$ is $\{0~\text{if}~i< 12~\text{else}~1\}$, while $M_2$ is obtained after toggling two zeros in $M_1$ to ones (for $i =4$ and $i=6$). It is evident that using $M_2$ transfers the shape of the faces more strongly. }
\label{fig:stymix}
\end{figure}

	In Fig.~\ref{fig:stymix}, we show how we can control the aspects of styles being transferred by choosing appropriate style-mixing mask $M$. We select two masks $M_1$ and $M_2$. $M_1$ is $\{0~\text{if}~i<12~\text{else}~1\}$, while $M_2$ is obtained after toggling two zeros in $M_1$ to ones (for $i =4$ and $i=6$). It is evident that using $M_2$ transfers the shape of the faces more strongly.
	
\begin{figure}[h]
\begin{center}
	\includegraphics[width=0.50\linewidth]{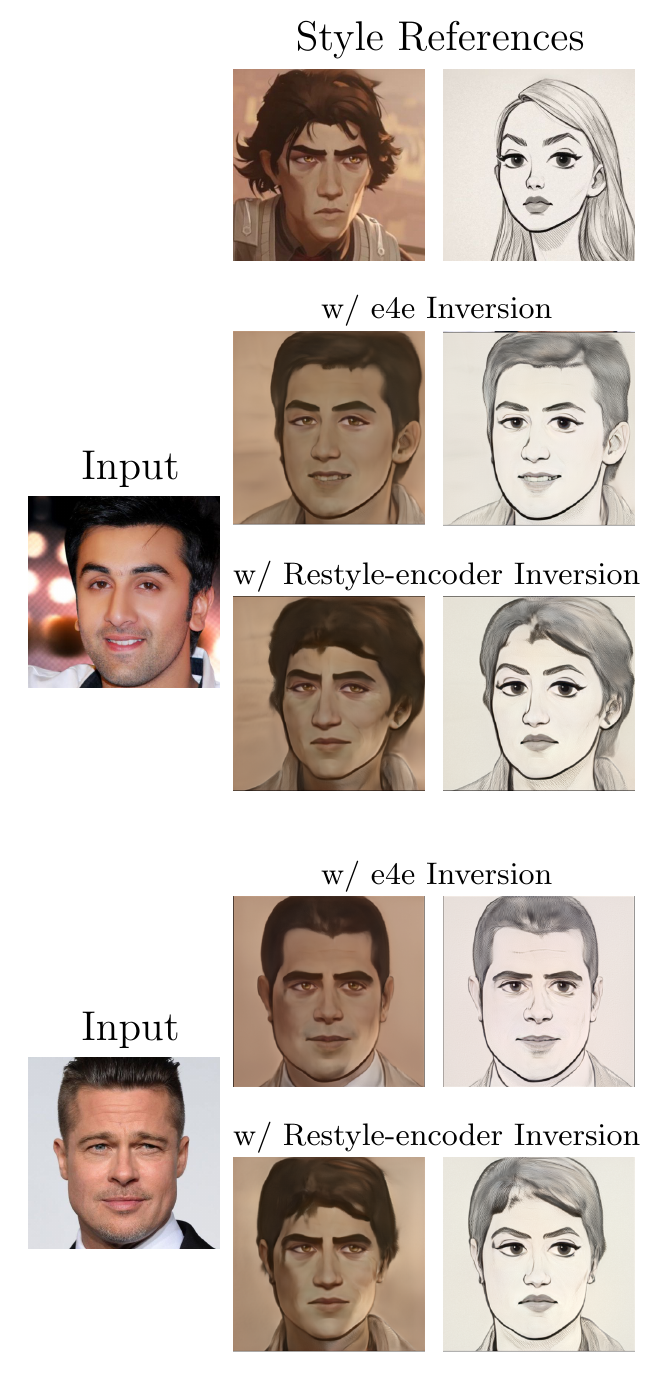}
\end{center}
\caption{\small{\textbf{Effect of GAN Inversion method.  }} Varying the GAN inversion method gives variations in stylizations as well. In this example, we compare between two inversion methods, e4e and Restyle-encoder. It can be seen that the outputs of Restyle-encoder transfers more style from style image to input image, including the expressions, thus the outputs take the expressions from the reference styles instead of inputs.}
\label{fig:inv}
\end{figure}

	In Fig.~\ref{fig:inv}, we show effect of changing GAN Inversion method from e4e to Restyle-encoder. Using the later transfers more style, and copies the expressions from the reference image.

\begin{figure*}[t]
\begin{center}
	\includegraphics[width=0.87\linewidth]{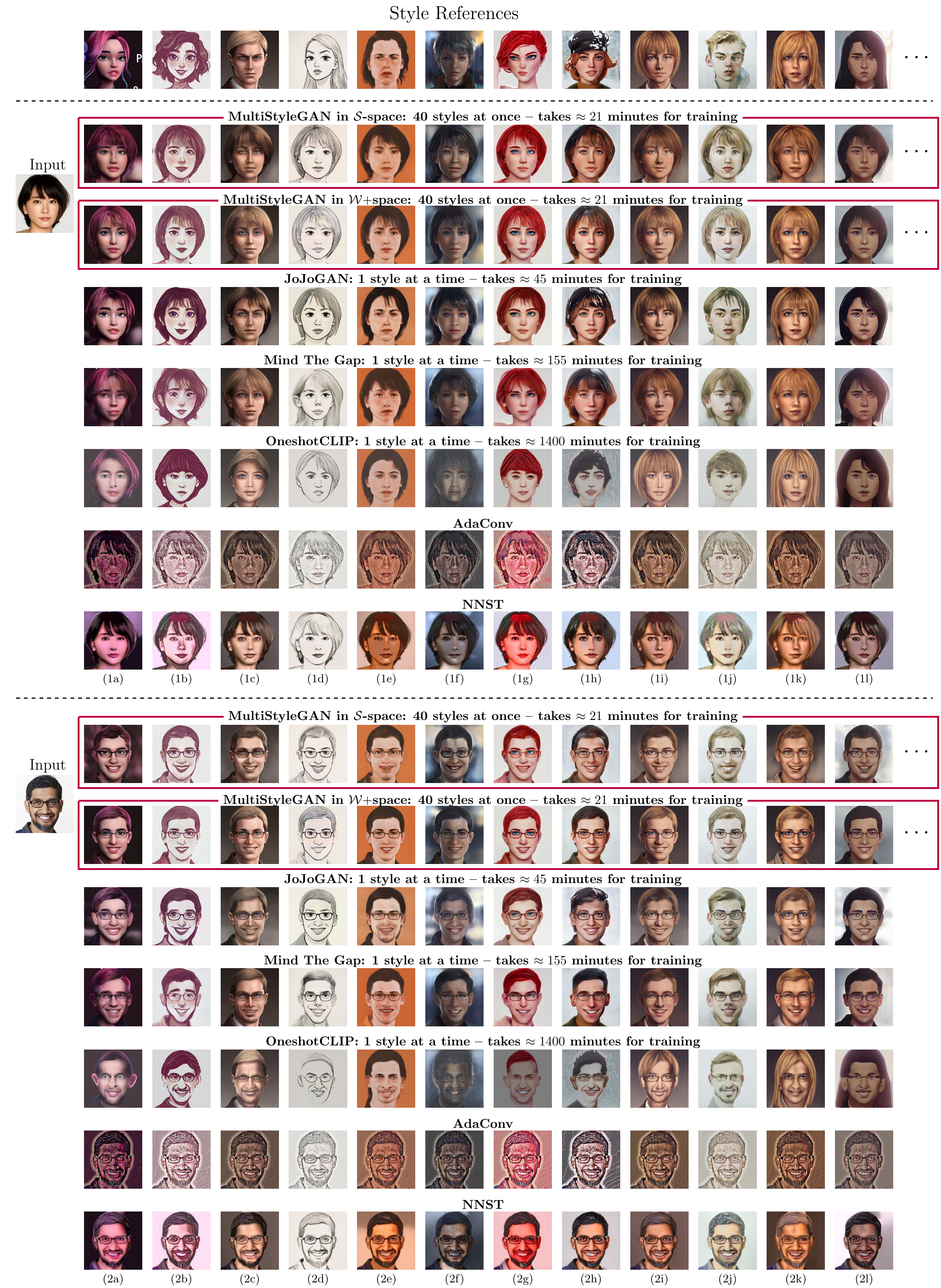}
\end{center}
\caption{\small{\textbf{Additional Results with Comparisons.} 
		Apart from producing $40$ stylizations at once, MultiStyleGAN brings noticeable improvements over current state-of-the-art methods. MultiStyleGAN prevents overfitting and preserves identity \& expressions of the inputs. JoJoGAN overfits on reference identity/expressions (see cols. 1c, 1j); MTG distorts the shape of face (see cols. 1d), chin (see cols. 1h), and nose (see col. 1j); OSC produces incomplete faces (see cols. 1h, 1j) and distorts the expressions (see cols. 2b, 2j); while the generic style transfer methods AdaConv and NNST fails to preserve the overall style itself. Fig. continues to next page.}}

\label{fig:compare_s}
\end{figure*}

\begin{figure*}[t]
\begin{center}
\includegraphics[width=0.87\linewidth]{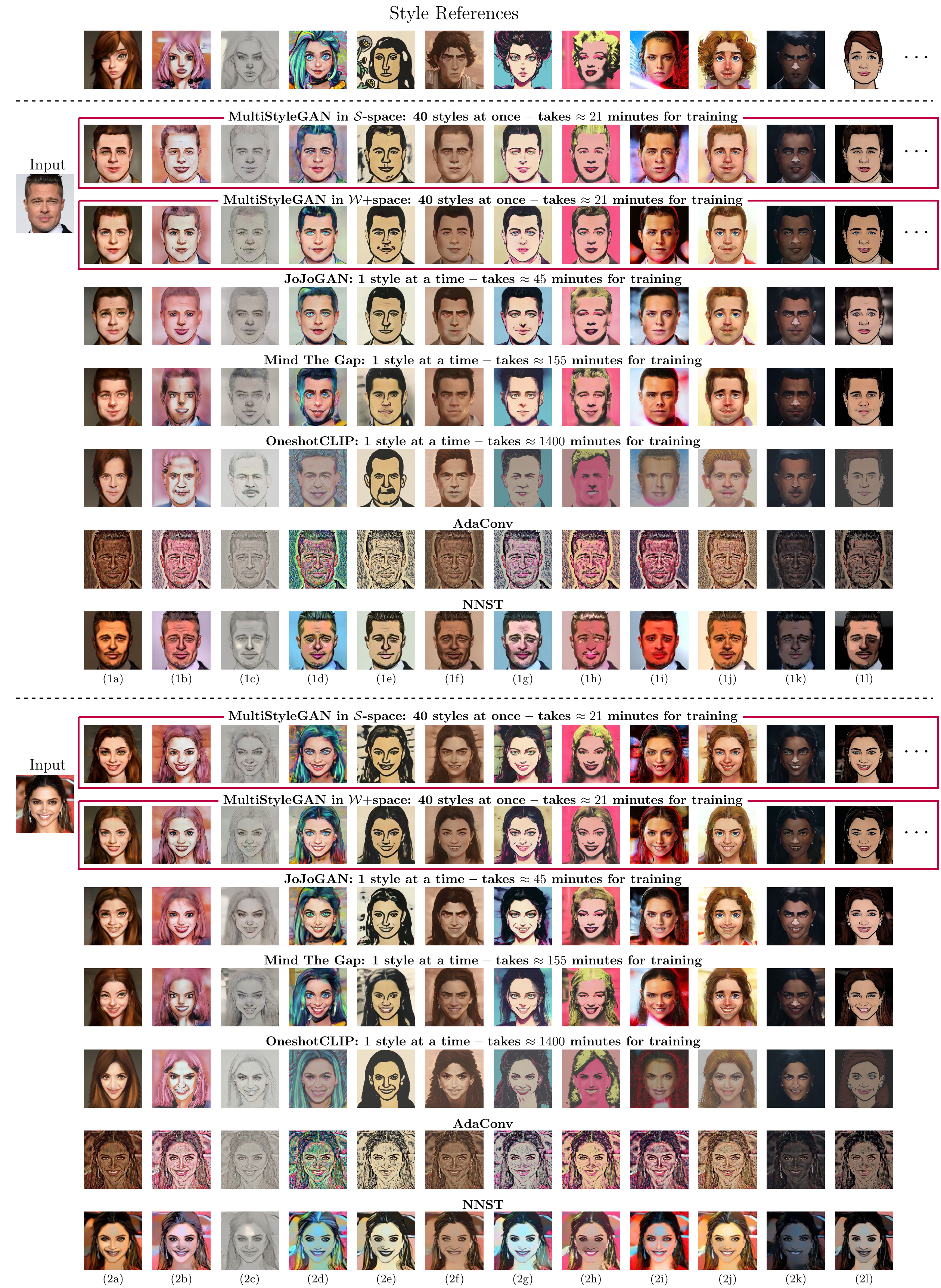}
\end{center}
\caption{\small{\textbf{Additional Results with Comparisons.} 
Apart from producing $40$ stylizations at once, MultiStyleGAN brings noticeable improvements over current state-of-the-art methods. MultiStyleGAN prevents overfitting and preserves identity \& expressions of the inputs. JoJoGAN overfits on reference identity/expressions (see cols. 1a, 2f); MTG distorts the shape of face (see cols. 1d), chin (see cols. 2g), and nose (see col. 1b); OSC produces incomplete faces (see cols. 2h, 1g) and distorts the expressions (see cols. 1e); while the generic style transfer methods AdaConv and NNST fails to preserve the overall style itself. Fig. continues to next page.}}

\label{fig:compare_s2}
\end{figure*}

\begin{figure*}[t]
\begin{center}
\includegraphics[width=0.87\linewidth]{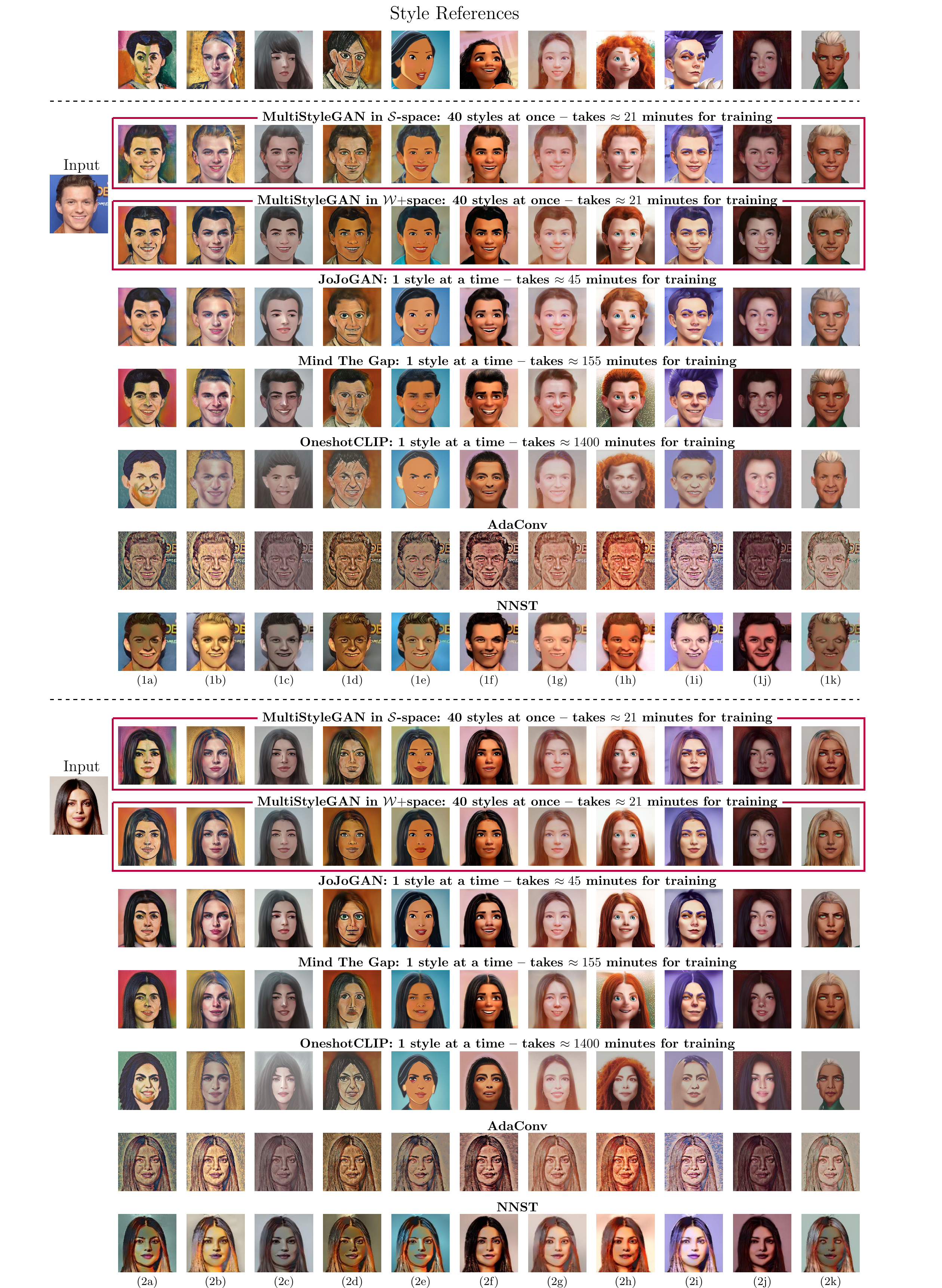}
\end{center}
\caption{\small{\textbf{Additional Results with Comparisons.} 
Apart from producing $40$ stylizations at once, MultiStyleGAN brings noticeable improvements over current state-of-the-art methods. MultiStyleGAN prevents overfitting and preserves identity \& expressions of the inputs. JoJoGAN overfits on reference identity/expressions (see cols. 1c, 2d); MTG distorts the shape of face (see cols. 2i), chin (see cols. 1i), and nose (see col. 2h); OSC produces incomplete faces (see cols. 1h, 2i) and distorts the expressions (see cols. 1e); while the generic style transfer methods AdaConv and NNST fails to preserve the overall style itself. }}

\label{fig:compare_s3}
\end{figure*}

	\noindent\textbf{A.4 Additional Results on Faces.}
	We provide additional results and comparisons to support our claims discussed in the main paper. In particular, Fig.~\ref{fig:compare_s}, Fig.~\ref{fig:compare_s2}, and Fig.~\ref{fig:compare_s3} provide additional results for comparisons on face images. Apart from providing stylizations on $35$ style references, we provide extensive set of comparisons with both the existing GAN-based one-shot stylization approaches (JoJoGAN, Mind The Gap, OneshotCLIP), and the generic style transfer methods AdaConv~\cite{Chandran2021AdaptiveCF} and NNST~\cite{Kolkin2022NeuralNS}. Additionally, we provided stylizations using the $\mathcal{W}+$ space variant of MultiStyleGAN as well. These figures contain diverse set of input images and reference styles - including some styles from JoJoGAN and Mind the Gap manuscripts as well. It is evident that the results are consistent with the qualitative and quantitative evidences provided in the main manuscript. 
	
\begin{figure*}[t]
\begin{center}
\begin{tabular}{c}
\includegraphics[width=0.98\linewidth]{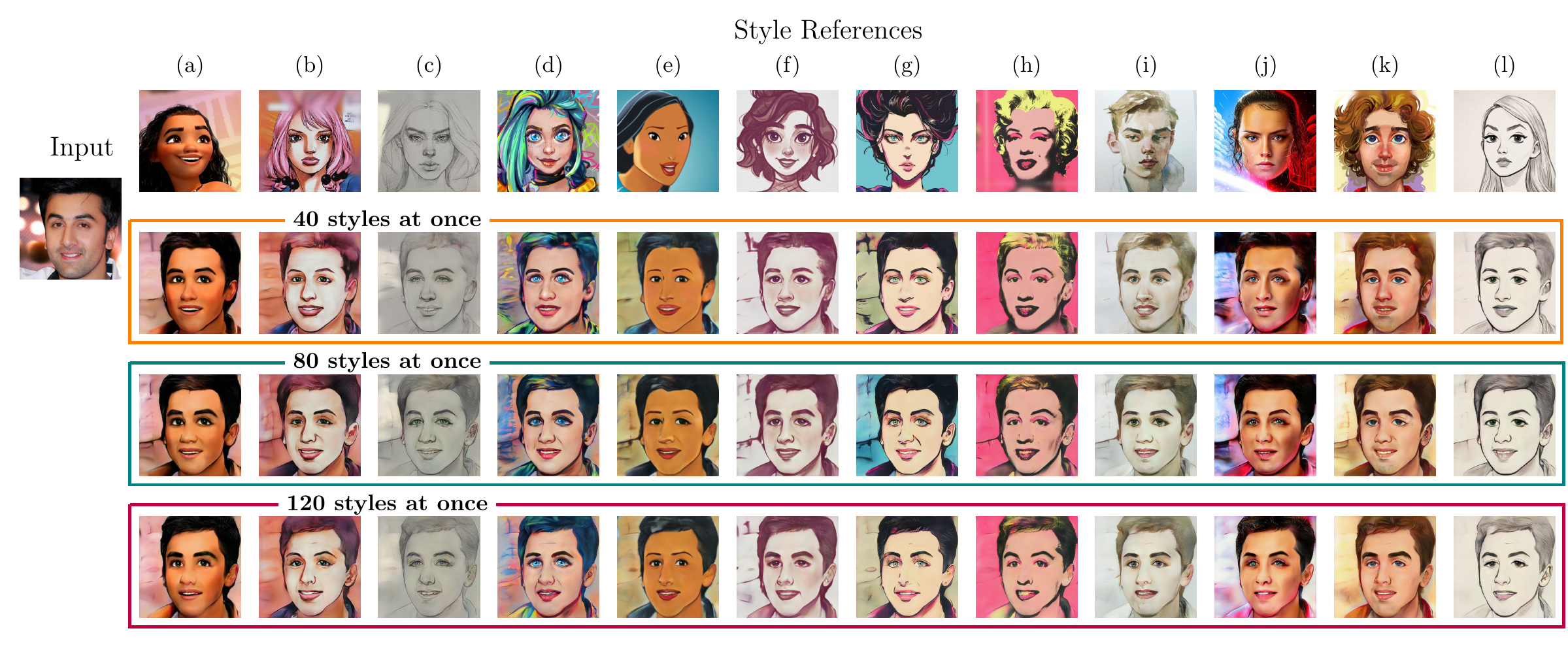}
\\ \addlinespace \addlinespace \addlinespace \addlinespace
\includegraphics[width=0.98\linewidth]{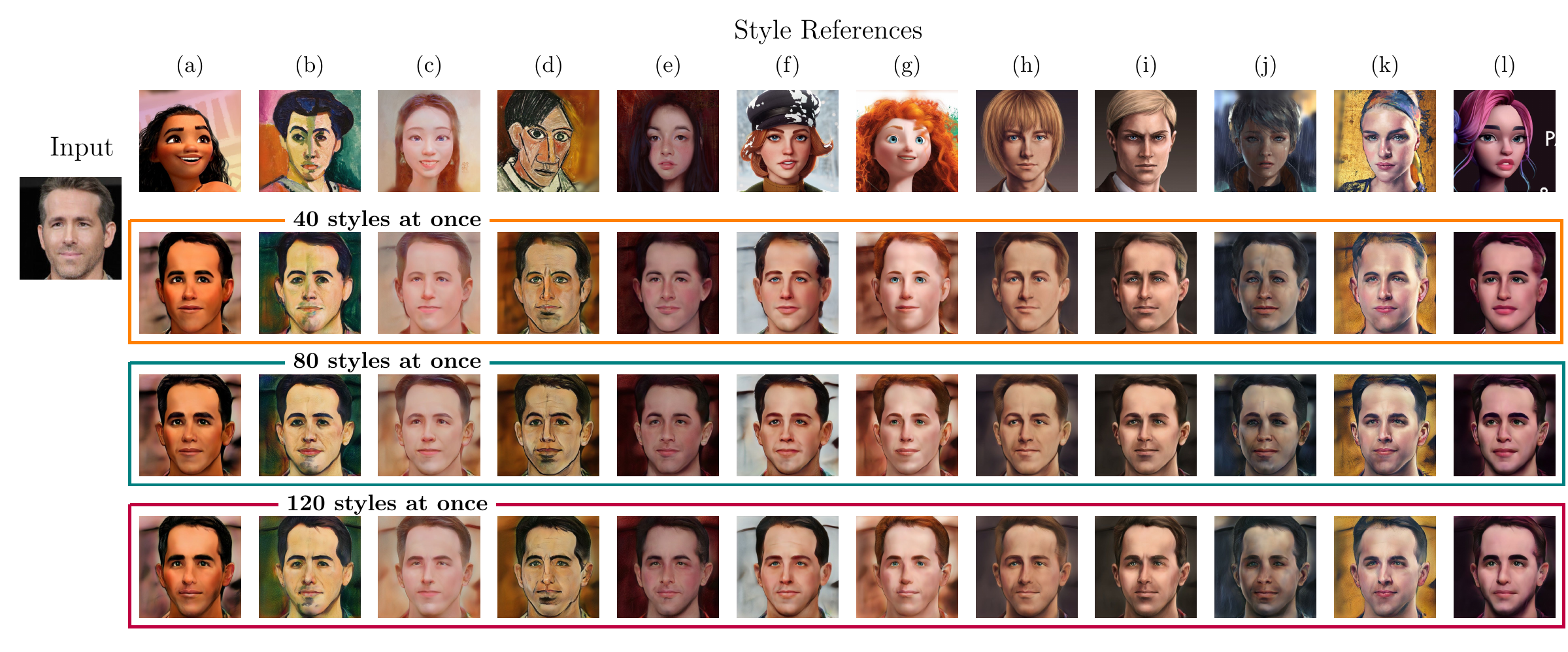}
\end{tabular}

\end{center}
\caption{\small{\textbf{Additional Results: Increasing total number of styles. }Barring some minor variations, quality of stylizations remain consistent for MultiStyleGAN as total number of styles $N$ are increased. We start with a set of $40$ styles and increase it to $N=80$ and $N=120$ for MultiStyleGAN model in $\mathcal{S}$-space. Stylizations for each case is depicted above for comparison.}}

\label{fig:no_of_styles_s}
\end{figure*}

	Further, Fig.~\ref{fig:no_of_styles_s} provides extended set of results for trade-off between MultiStyleGAN's performance and number of reference styles $N$ being used during training. We compare the stylizations obtained using $N=40$, $N=80$, and $N=120$ with MultiStyleGAN in $\mathcal{S}$-space. It is evident that our method is able to maintain the stylization performance as value of $N$ increases. 
	
\begin{figure*}[t]
\begin{center}
\includegraphics[width=0.65\linewidth]{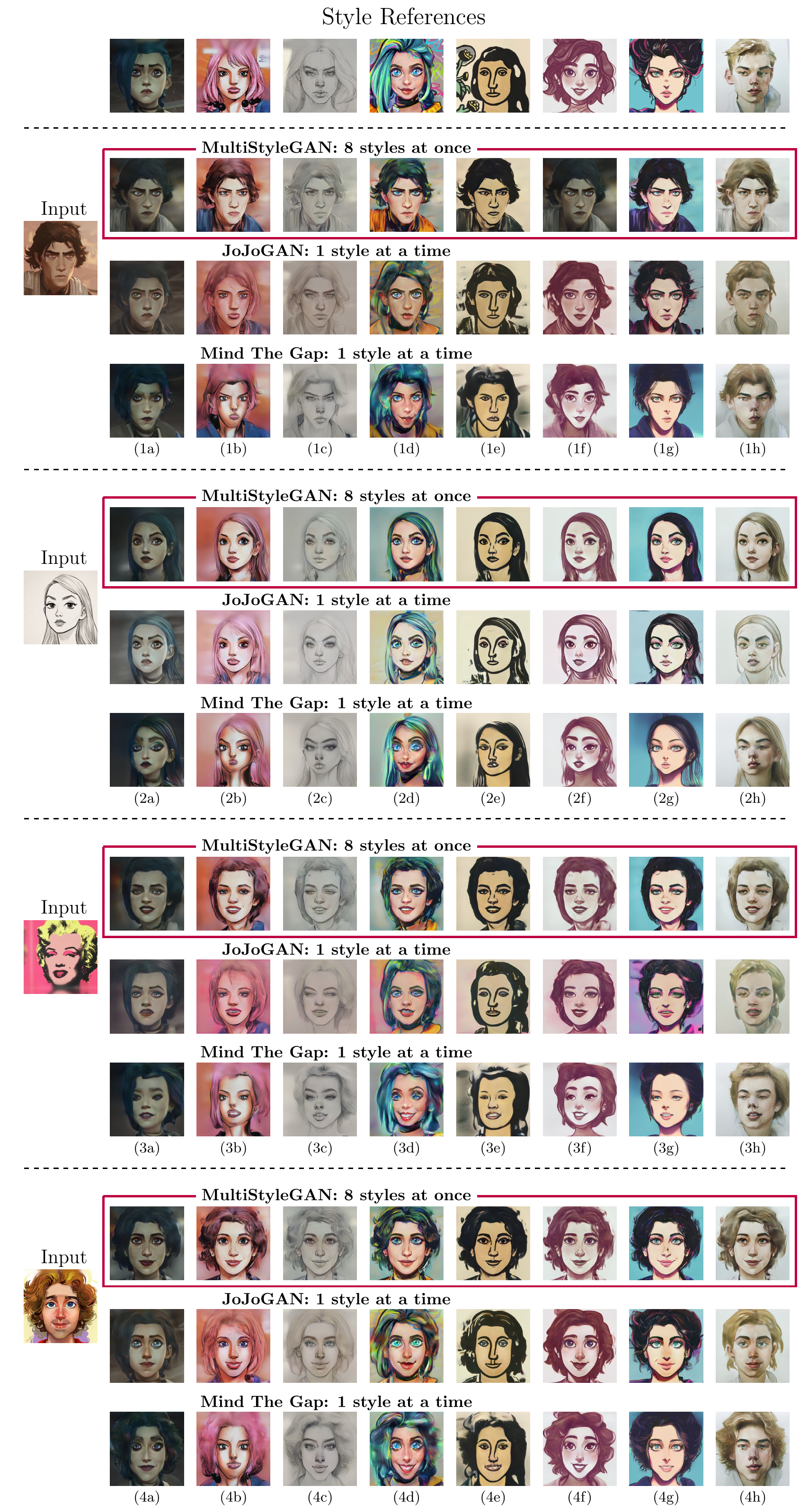}
\end{center}
\caption{\small{\textbf{Stylizing reference image itself.}} Our model can stylize out-of-domain inputs. We use unseen style images as inputs, and our model is able to stylize it to the reference style while preserving the identity. JoJoGAN and Mind the Gap clearly fails for such out-of-domain examples due to overfitting and shape distortions respectively.}

\label{fig:cross_s}
\end{figure*}

\begin{figure*}[t]
\begin{center}
\includegraphics[width=0.95\linewidth]{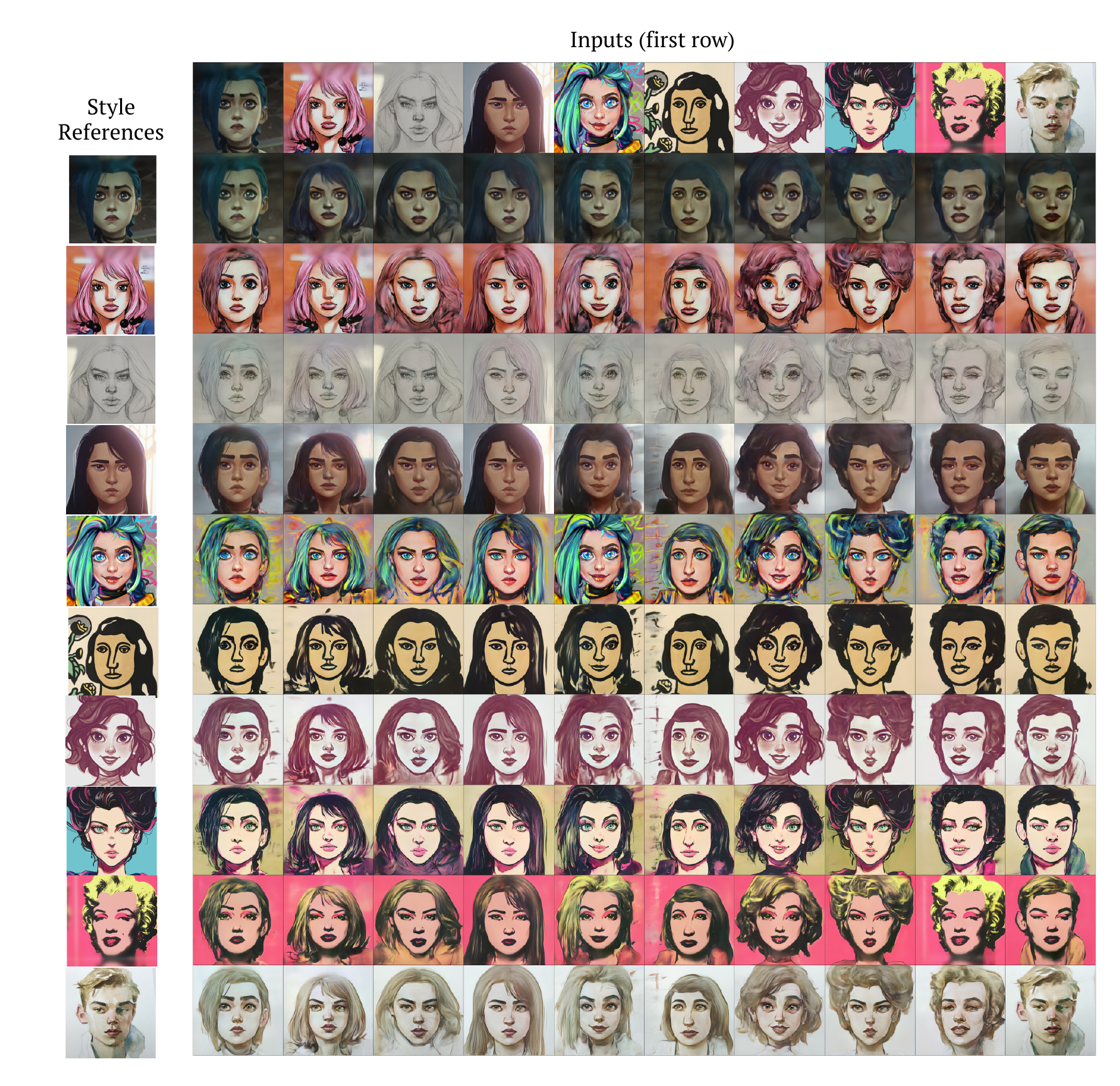}
\end{center}
\caption{\textbf{Additional Results: Stylizing reference image itself.} In this example, the style references used for training are passed as the inputs to our MultiStyleGAN for the task of re-stylization. Re-stylization results are shown in the form of matrix, where the first column on left represents the inputs, and the first row on top represents the reference style.} 

\label{fig:cross_matrix}
\end{figure*}

	Fig.~\ref{fig:cross_s} and Fig.~\ref{fig:cross_matrix} provide additional results and comparisons for re-stylization of reference styles, i.e., stylizing out-of-domain inputs that are not realistic images.	

\begin{figure*}[t]
\begin{center}
\includegraphics[width=0.92\linewidth]{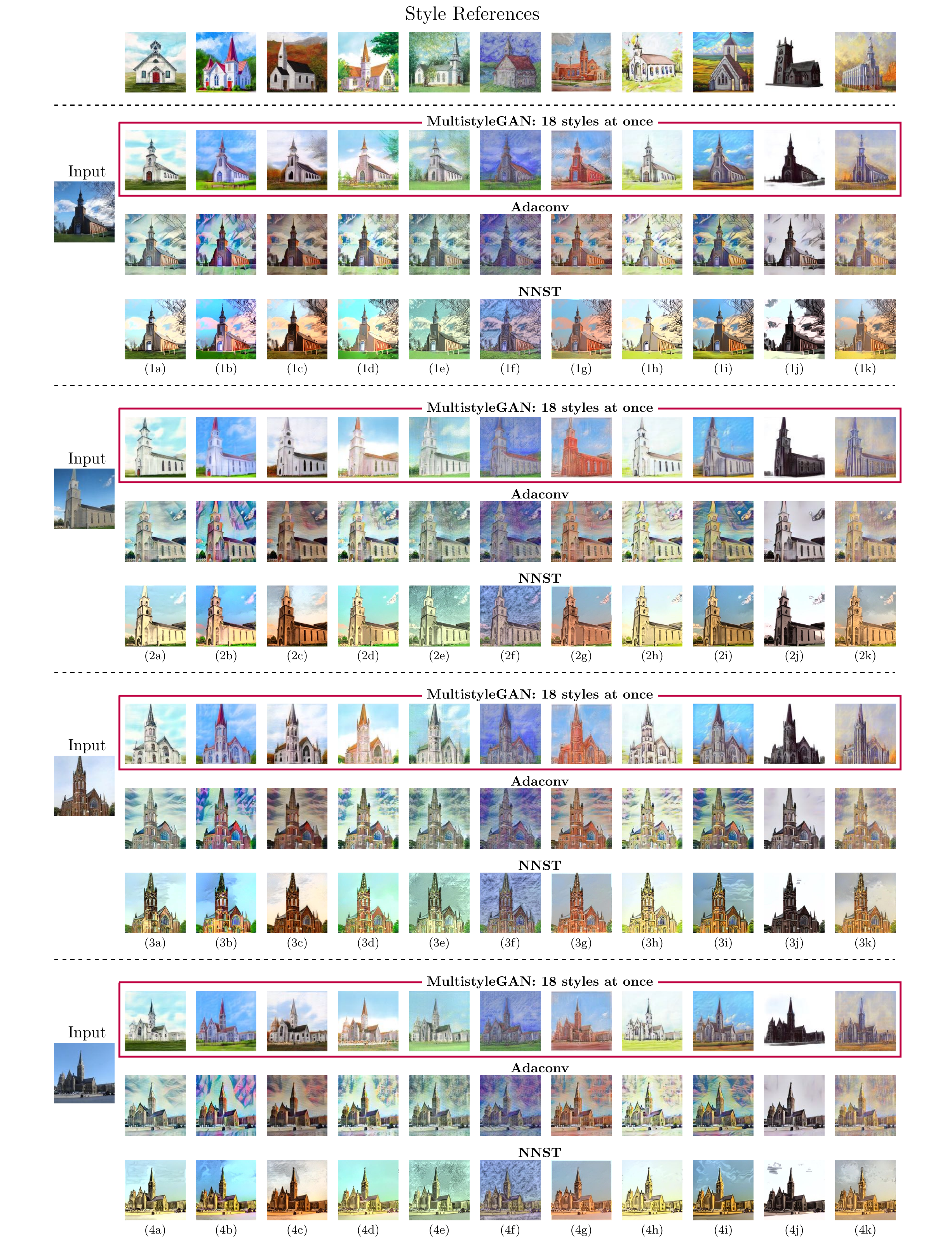}
\end{center}
\caption{{\small \textbf{Additional on Non-face Images. } MultiStyleGAN can successfully stylize non-face inputs such as churches outperforming other generic style transfer methods. The above results are obtained with MultistyleGAN model in $\mathcal{S}$-space with $N=18$.}}
\label{fig:church_s}
\end{figure*}

\begin{figure*}[t]
\begin{center}
\includegraphics[width=0.92\linewidth]{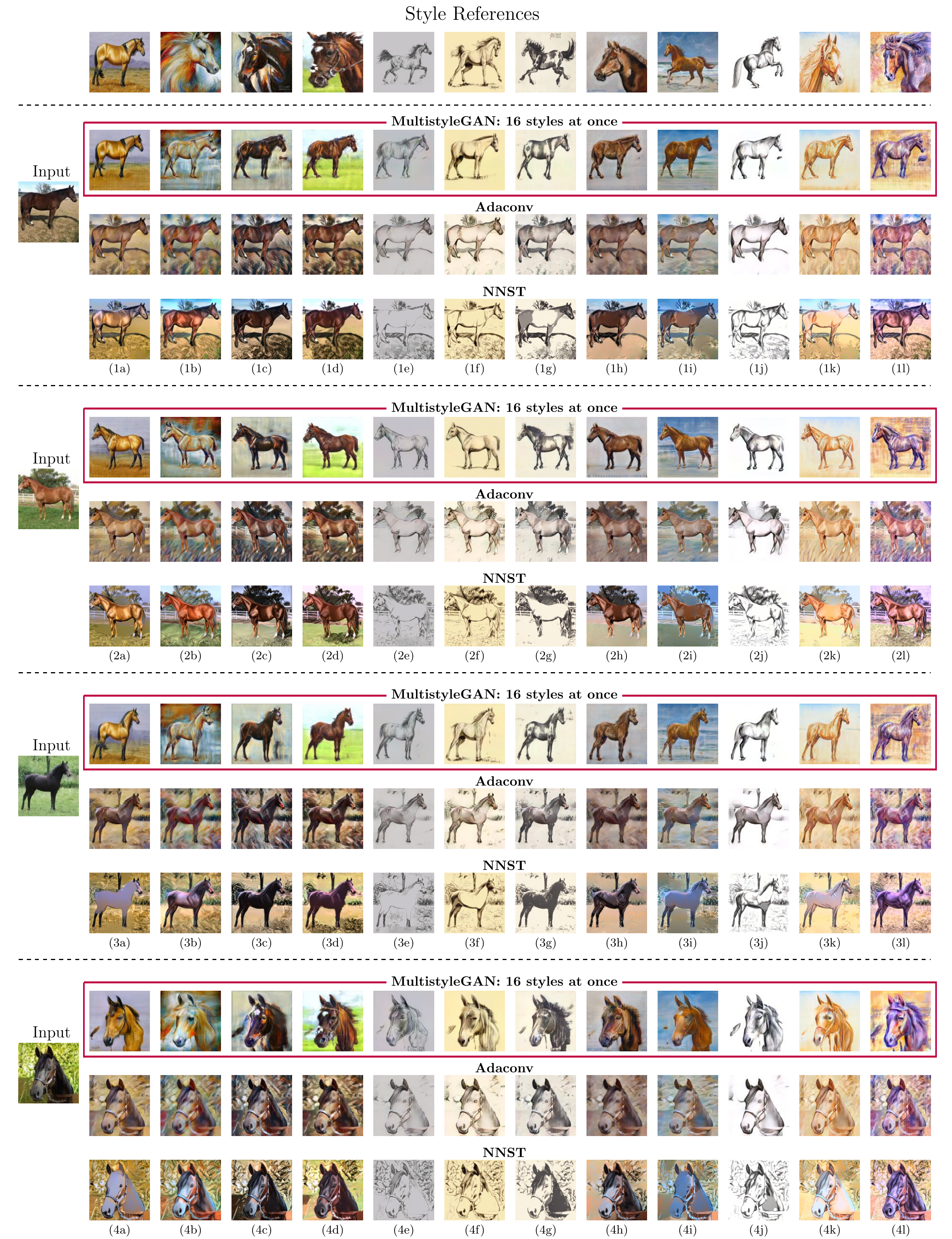}
\end{center}
\caption{{\small \textbf{Additional on Non-face Images. } MultiStyleGAN can successfully stylize non-face inputs such as horses outperforming other generic style transfer methods. The above results are obtained with MultistyleGAN model in $\mathcal{S}$-space with $N=16$.}}
\label{fig:horse_s}
\end{figure*}

\begin{figure*}[t]
\begin{center}
\includegraphics[width=0.97\linewidth]{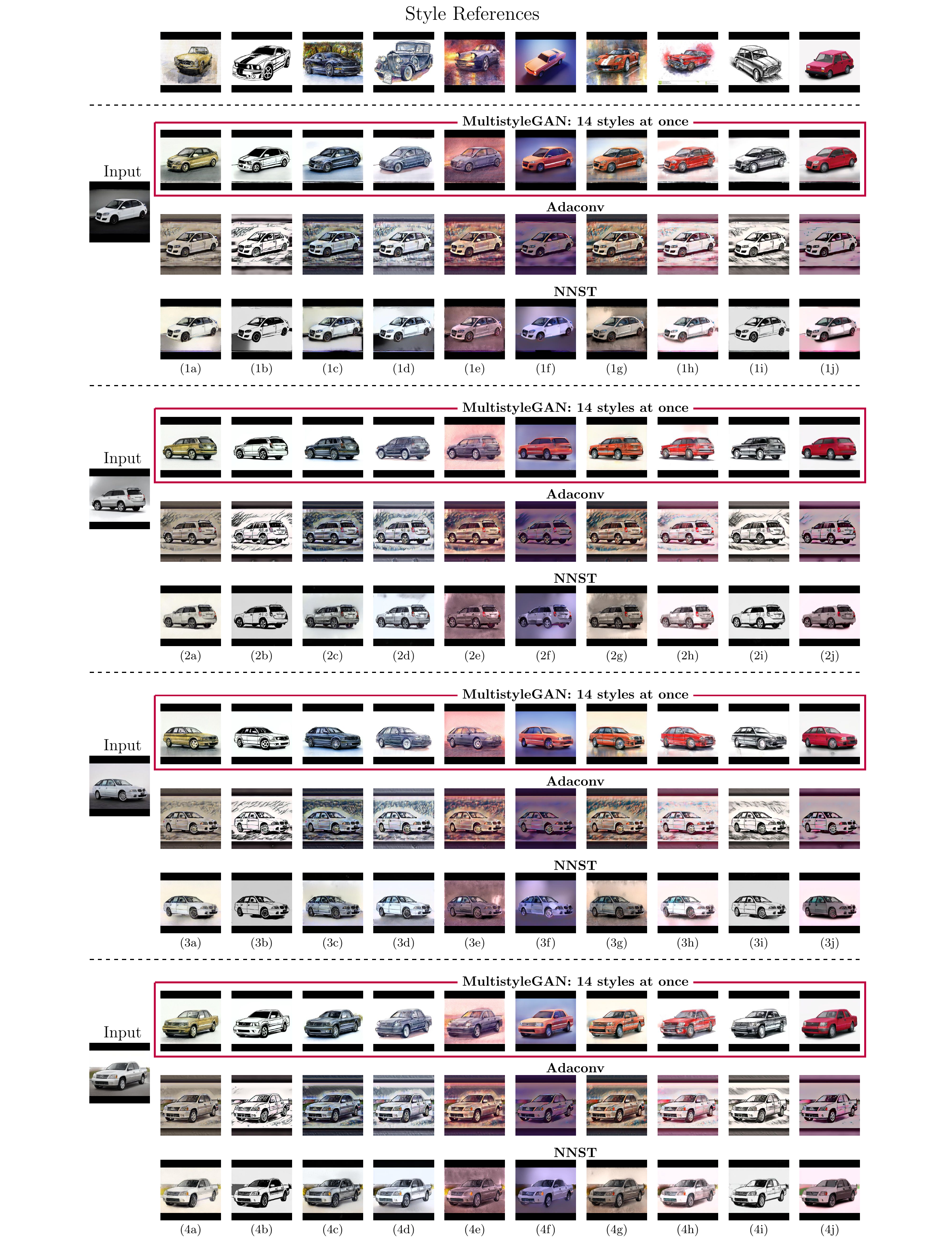}
\end{center}
\caption{{\small \textbf{Additional on Non-face Images. } MultiStyleGAN can successfully stylize non-face inputs such as cars outperforming other generic style transfer methods. The above results are obtained with MultistyleGAN model in $\mathcal{S}$-space with $N=14$.}}
\label{fig:car_s}
\end{figure*}

	\vspace{0.5em}
	\noindent\textbf{A.5 Additional Results on Churches, Horses, and Cars. }We also provide extensive set of stylization results and comparisons on non-face images such as churches (Fig.~\ref{fig:church_s}), horses (Fig.~\ref{fig:horse_s}), and cars (Fig.~\ref{fig:car_s}). In the case of churches we use $N=18$, while we use $N=16$ and $N=14$ for horses and cars respectively. We also provide comparisons of our results with state-of-the-art generic style transfer approaches AdaConv~\cite{Chandran2021AdaptiveCF} and NNST~\cite{Kolkin2022NeuralNS}.  Results indicate that AdaConv and NNST fail to reproduce the correct colorization in their outputs and lack semantic understanding of the scene, while our method generates faithful stylizations while preserving the semantic structure of the input for all three types of images.
	
\begin{figure*}[t]
\begin{center}
\includegraphics[width=0.70\linewidth]{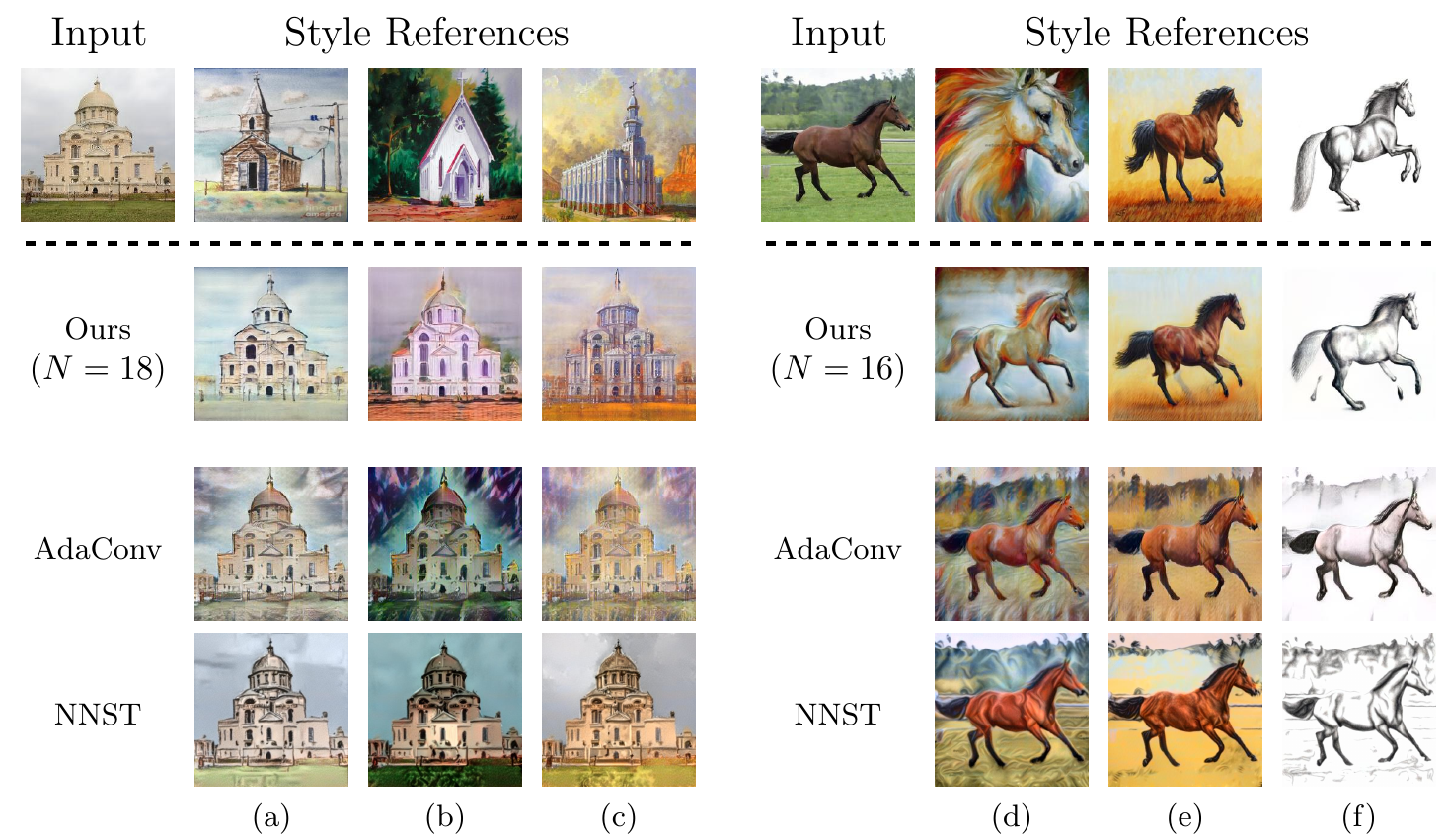}
\end{center}
\caption{\small{\textbf{Failure cases on non-face images. }While our model remains perceptually and qualitatively superior to AdaConv and NNST, it can encounter failure cases pertinent to both the style and content preservation. In the case of churches, the color of the style reference can get washed out (see col. a), or the structure of the input image does not transfer to the output (see col. c). For horses, although the stylizations retain expected quality, the pose of the input horse image changes in the stylized image (col. d, e) and the legs of the horse is not reconstructed fully (col. f) due to poor quality of GAN inversion.}  }
\label{fig:non_face_failure}
\end{figure*}

	We also include a set of failure cases on Church and Horse images in Fig.~\ref{fig:non_face_failure}. While MultistyleGAN remains perceptually and qualitatively superior to AdaConv and NNST, it encounters failure cases pertinent to both the style and content preservation. For example, in the case of churches, the color of the style reference can get washed out (see col. a), or the structure of the style reference does not transfer to the input image (see col. c) leading to distortions. For horses, we highlight issues with the GAN inversion which is a key intermediate step in our method. Although the stylizations retain expected quality, the pose of the input horse image changes in the stylized image as the encoder used for inversion fails to provide a faithful recovery. Adaconv and NNST do not encounter such issues in maintaining the structure of the input image, however, their quality of stylization remains poor.

\newpage

\newpage

\newpage
\newpage

\newpage

\newpage

\newpage

\newpage

\end{document}